%% file: tro.tex
\documentclass[lettersize,journal]{IEEEtran}
\usepackage{amsmath,amssymb,amsfonts, amsthm}
\usepackage{algorithmic}
\usepackage{graphicx}
\usepackage{textcomp}
\usepackage{pifont}
\usepackage{lipsum}
\usepackage{multirow, booktabs, makecell}
\usepackage{color, colortbl}
\usepackage{url}
\usepackage{subcaption}
\usepackage{float}
\usepackage{nccmath}
\usepackage{hhline}
\usepackage{setspace}
\usepackage{bbm}
\usepackage{threeparttable}
\usepackage[capitalize]{cleveref}
\usepackage[sorting = none, backend = bibtex, style=numeric-comp]{biblatex}
\AtBeginBibliography{\small}

\usepackage{tikz}
\usepackage{pgfplots}
\pgfplotsset{compat=1.16} 

\crefname{section}{Sec.}{Secs.}
\Crefname{section}{Section}{Sections}
\Crefname{table}{Table}{Tables}

\newcommand{\cmark}{\ding{51}}%
\newcommand{\xmark}{\ding{55}}%
\newcommand{\ts}{\textsuperscript}

\DeclareMathOperator{\arctantwo}{arctan2}

\begin{document}

\title{HALS: A Height-Aware Lidar Super-Resolution Framework for Autonomous Driving}

\author{George Eskandar$^{1}$,
Sanjeev Sudarsan$^{1}$,
Karim Guirguis$^{2}$,
Janaranjani Palaniswamy$^{1}$,
Bharath Somashekar$^{1}$,
Bin Yang$^{1}$ \\
\thanks{The research leading to these results is funded by the German Federal Ministry for Economic Affairs and Energy within the project "AI Delta Learning".  The authors would like to thank the consortium for the successful cooperation.} 
{$^1$ University of Stuttgart, Institute of Signal Processing and System Theory, Stuttgart, Germany   \\
$^2$ Robert Bosch GmbH, Renningen, Germany}
}




\maketitle

\begin{abstract}
Lidar sensors are costly yet critical for understanding the 3D environment in autonomous driving. High-resolution sensors provide more details about the surroundings because they contain more vertical beams, but they come at a much higher cost, limiting their inclusion in autonomous vehicles. Upsampling lidar pointclouds is a promising approach to gain the benefits of high resolution while maintaining an affordable cost. Although there exist many pointcloud upsampling frameworks, a consistent comparison of these works against each other on the same dataset using unified metrics is still missing. In the first part of this paper, we propose to benchmark existing methods on the Kitti dataset. In the second part, we introduce a novel lidar upsampling model, HALS: Height-Aware Lidar Super-resolution. HALS exploits the observation that lidar scans exhibit a height-aware range distribution and adopts a generator architecture with multiple upsampling branches of different receptive fields. HALS regresses polar coordinates instead of spherical coordinates and uses a surface-normal loss. Extensive experiments show that HALS achieves state-of-the-art performance on 3 real-world lidar datasets. 
\end{abstract}



\section{Introduction}
\label{sec:intro}

Light detection and ranging (lidar) pointclouds are vital for the geometrical understanding of a surrounding environment for autonomous vehicles. Numerous tasks are dependent on 3D information from lidar, such as 3D object detection~\cite{pvrcnn, pointpillars, second, rangercnn, sessd}, 3D semantic segmentation~\cite{rangenet++}, localization~\cite{localization}, mapping~\cite{mapping} and path planning~\cite{trajectory}. Mounted on the vehicle, lidar emits pulses of infrared light waves to retrieve accurate 3D position information. Although the horizontal resolution of lidar is high, the vertical resolution is usually low and depends on the number of channels present in the sensor (16, 32, 64 and 128 typically). Denser pointclouds in the vertical direction are desirable because they provide more cues about the environment, thereby improving the performance of many computer vision tasks~\cite{richter2022understanding}. However, high-resolution lidar sensors come at a significantly higher cost~\cite{unetlidar}, which would impede their commercial use in autonomous driving. It is often the case that high-resolution sensors are used in test drives to validate a perception pipeline in an autonomous vehicle, but are then replaced by cheaper low-resolution sensors for commercial purposes.  

\input{Figures/teaser}
Lidar upsampling (or super-resolution) is a promising approach toward achieving a trade-off between cost and performance. Compared to image super-resolution, lidar super-resolution is more challenging because the pointclouds are not structured in a grid like images. Moreover, neural networks that handle 3D data are computationally expensive and might only scale to a small number of points. Although many pointcloud super-resolution algorithms have achieved remarkable performance, two main challenges prevent us from assessing which methods work better for lidar in autonomous driving. First, there has yet to be a consistent benchmark to compare the different pointcloud upsampling algorithms. For instance,~\cite{punet, pugan, patchupsample, argcn} were tested on datasets containing synthetic objects with fewer points ($\sim10k$) than a typical lidar scan ($\sim100k$). \cite{unetlidar, iln} were designed and trained on lidar pointclouds extracted from a driving simulator like Carla~\cite{carla}, while others have been evaluated on real datasets~\cite{uqlidar, swd, lidarsrresnet, unetlidar}. The second challenge is that different evaluation metrics are used for various models. 

We argue that benchmarking the different approaches on the same dataset with unified metrics would help identify their strengths and shortcomings while highlighting the best practices in this emerging field. It would also pave the way toward developing better generative models for lidar upsampling. Thus, this work attempts to close the gap on two questions: \textit{how to benchmark and evaluate the different pointcloud super-resolution methods on real lidar scans?} And \textit{how to upsample real lidar pointclouds while preserving the 3D geometry of the environment?} Most importantly, we only focus on real-world lidar data because methods that work well on synthetic data might not generalize well to real data~\cite{syn2reallidar}.

\noindent\textbf{Contribution.} This work is organized into 2 parts. In the first part, we present a consistent comparative evaluation of several representative frameworks on the Kitti Raw dataset~\cite{kitti}. We classify the methodologies into 2 groups: point-based and grid-based methods. The former use PointNet operations directly on the pointcloud, while the latter project the pointcloud on image coordinates and deploy 2D convolutional networks. Our key insight from the benchmark is that grid-based methods are superior than point-based ones because of their vertically large receptive field which spans multiple lines. In the second part of this work, we take a closer look at the geometry of lidar scans and find that real-world lidar range image exhibits a height-dependent range distribution. Specifically, the beams corresponding to high elevation angles (located in the upper part of the range image) have a higher average range and standard deviation. This insight is often overlooked in previous works, leading to inaccurate shapes or noisy points in the generated pointclouds. Motivated by these findings, we introduce HALS, a Height-Aware Lidar Super-resolution framework that achieves state-of-the-art results on 3 real-world lidar datasets: Kitti Raw~\cite{kitti}, Kitti Object~\cite{kitti} and Nuscenes~\cite{nuscenes}. The proposed approach is designed to best match the range distribution of the high-resolution ground truth (\cref{fig:teaser}). First, we design a generator with 2 upsampling branches of different receptive fields and fuse the 2 outputs using confidence maps that model the uncertainty of each branch in its prediction. The multi-scale receptive fields allow the network to adapt to the height-dependent range distribution by knowing \textit{where to gather information}. Moreover, instead of regressing spherical coordinates, we regress the polar range and height of the pointcloud, achieving a more accurate synthesis by reducing the vertical quantization error in the 2D-to-3D projection. Finally, we show how to use the surface normal of the generated scene to improve the lidar scan.   

We summarize our contributions as follows:
\begin{itemize}
    \item We present a strong empirical evaluation of previous works on the Kitti Raw dataset with unified metrics, which reveals the superiority of grid-based methods relative to point-based methods. The main reason for this is a vertically larger receptive field enabled by a structured input representation.
    \item We propose a novel grid-based generator architecture that upsamples the pointcloud at different receptive fields, so that it can adapt to the range distribution of the upper and lower parts of the range image. Each upsampling branch outputs a confidence map of its own prediction. 
    \item Upon observing that vertical lidar channels are not uniformly spaced, we propose to use polar coordinates in the input and output range image. Moreover, we show the importance of adopting a surface normal loss besides the conventional $\mathcal{L}_1$-loss to preserve structural details.
    \item Evaluations on three standard datasets, Kitti Raw, Kitti Object detection~\cite{kitti} and Nuscenes~\cite{nuscenes}, show the effectiveness and superior performance of HALS compared to the baselines. 
\end{itemize}

\section{Related Works}
\label{sec:rw}
\noindent
\textbf{Image Super-Resolution.} Image super-resolution has thrived in recent years thanks to advancements in deep learning. A pioneering work, SRCNN~\cite{srcnn}, developed a CNN-based architecture to upsample images. Then, a series of enhanced frameworks have been proposed~\cite{srgan, esrgan, swinir}. For instance, SRResNet~\cite{srgan} is a framework which deploys residual blocks and a transposed convolution layer to improve visual perception. A state-of-the-art method SWIN-IR replaces the SRResNet backbone with the SWIN transformer~\cite{swin}.

\noindent
\textbf{Generative Models for Lidar.}  Recently, there has been a growing interest in developing generative models for lidar \cite{deepgm}. A similar line of work, namely sequential lidar prediction \cite{spf2, monet, 3dsptemporal, slpc}, aims to improve the trajectory forecasting task by generating future lidar pointclouds. Most of these works synthesize pointclouds by generating their 2D spherical projection, also known as range image. The projection is mathematically defined in \cref{sec:overview}. The networks are trained by minimizing an $\mathcal{L}_1$-loss with respect to the ground truth. Other works~\cite{ spf2,3dsptemporal} minimize the chamfer distance additionally.

\noindent
\textbf{Pointcloud Upsampling} or super-resolution refers to the task of adding more points to a pointcloud while preserving its shape. Similar to image super-resolution, most recent pointcloud super-resolution algorithms~\cite{punet, patchupsample, pugan, argcn} are based on deep learning. The learning occurs in a self-supervised way, by downsampling the pointclouds using either a uniform or non-uniform point dropout scheme, then upsampling them with deep learning using a reconstruction loss. Most of the works in this task focused on (but are not limited to) upsampling pointclouds of single objects as opposed to pointclouds of whole scenes. The question of whether these methods can extend to pointclouds depicting a whole scene is yet to be answered. Note that in the case of single-object pointclouds, even if non-uniform downsampling is used to get the low-resolution input, the high-resolution ground truth is spatially uniform (the point density is more or less consistent across the object). Architectures of these works~\cite{punet, patchupsample, pugan} are based on PointNet-operations~\cite{pointnet}, while AR-GCN~\cite{argcn} uses graph convolution layers. Since these methods operate on the raw points in 3D, they are commonly referred to as \textit{point-based methods}.



\noindent
\textbf{Lidar Upsampling} refers to the task of upsampling a low-resolution lidar pointcloud, which typically originates from a lidar sensor with a few number of vertical channels. It is a sub-task of the more general and previously discussed pointcloud upsampling task. Lidar pointclouds in autonomous driving depict a whole scene with many objects of different classes. These pointclouds have in general a non-uniform spatial distribution and contain a considerably larger number of points than single objects pointclouds ($100k$ vs. $10k$). 

A few recent works~\cite{lidarsrresnet,unetlidar, iln, swd, sga, uqlidar} have explicitly addressed this task. LIDAR-CNN \cite{lidarsrresnet} deploys a CNN~\cite{srgan}-based architecture and an $\mathcal{L}_1$-loss function. An additional point segmentation network is used to guide the generator to a better synthesis through a feature matching loss, but requires additional labeling costs. LIDAR-SR~\cite{unetlidar} deploys a Bayesian UNet-based architecture to prevent overfitting and reduce the number of noisy points. ILN~\cite{iln} uses an implicit neural architecture which generates interpolating weights for each point's nearest neighbours (in the range image). Both ILN and LIDAR-SR were trained on synthetic datasets generated from the Carla~\cite{carla} simulator. Similar to LIDAR-SR, \cite{uqlidar} designs a grid-based method with uncertainty quantification that outperforms LIDAR-SR in speed but shows a comparable upsampling performance. These 4 methods~\cite{lidarsrresnet, unetlidar, iln, uqlidar} project the lidar scan on a 2D plane and use a CNN to upsample the lidar scan in the image space. We refer to them as \textit{grid-based} methods, because the lidar pointcloud are structured in a 2D grid after the spherical projection. 

\textit{Point-based methods} which were developped for generic pointcloud upsampling (discussed in the previous paragraph~\cite{punet, pugan, patchupsample, argcn}) can also be applied to lidar pointclouds, although this has not yet been studied. Additionally, two point-based methods have been recently introduced to explicitly address lidar upsampling: SWD~\cite{swd} employs an architecture with edge convolutions and a sliced Wasserstein distance loss, while \cite{sga} proposes a downsampling algorithm without training and shows a comparable performance to LIDAR-SR~\cite{unetlidar} on real outdoor scenes. We summarize this section in \cref{fig:rw}.  

\input{Figures/rw}
\input{Figures/results}

\section{Overview of upsampling algorithms}
\label{sec:overview}
In this section, we explore how point-based methods and grid-based methods perform in the lidar upsampling task by conducting a consistent comparison on the Kitti Raw dataset~\cite{kitti} with unified metrics.

\noindent\textbf{Problem Formulation.} Formally, let $\mathcal{P}$ be the input lidar pointcloud with $N$ points, where $\mathcal{P} = \{(x_i,y_i,z_i)\ : i=1,..,N\}$, and $z$ is the vertical height. $\mathcal{P}$ can be projected onto 2D image coordinates $(u,v)$ resulting in a range image representation $\mathcal{Q}$. We define $(H,W)$ to be the height and width of $\mathcal{Q}$, $f=f_{up} + f_{down}$ the vertical field-of-view of the lidar sensor, $r = \sqrt{x^{2} + y^{2} + z^{2}}$ the range of the point and $d = \sqrt{x^{2} + y^{2}}$ the radial distance. The 2D spherical projection can be expressed in \cref{eq:rangeimage} as follows:
\begin{equation}
\label{eq:rangeimage}
\begin{aligned}
    \small
    u &= \frac{1}{2}[1 - (\arctantwo(\frac{y}{x})\pi^{-1}]W, u \in [0,W] \\
    v &= [1-(\arcsin(zr^{-1}) + f_{up})f^{-1}]H, v \in [0,H] \\
\end{aligned}
\end{equation}
$\mathcal{Q}$ is commonly represented in spherical coordinates ($\mathcal{Q}_{u,v} = r$) \cite{rangenet++, rangercnn, unetlidar, lidarsrresnet, lidarcircular}. $u$ represents the azimuth while $v$ represents the elevation. In \cref{fig:teaser}, we illustrate a range image: bright values represent far away objects, while darker colors represent a small range value (near objects). Note that during the projection, multiple points can fall in the same bin $(u,v)$. In this case, we take the point with the smallest range, while other points are considered as occluded. On the other hand, there exists some bins with no point, these are empty bins for which $\mathcal{Q}_{u,v} = 0$. The number of empty bins is small but not negligible. For example, in the Kitti Object dataset, we find $15\%$ to $25\%$ of the total number of bins to be empty. \cref{eq:rangeimage} can be inverted to project the range image back into 3D space.  

We seek to transform a low-resolution pointcloud, $\mathcal{P}_{LR}$ to a high-resolution one, $\hat{\mathcal{P}}_{HR}$, with a larger number of points in the vertical direction. Note that both $\mathcal{Q}_{LR}$ and $\mathcal{Q}_{HR}$ have the same vertical field-of-view $f$, but different number of lines. This means that the upsampled range image, $\hat{\mathcal{Q}}_{HR}$, should have a bigger height $H$ and the same $W$ as $\mathcal{Q}_{LR}$. In practice, we are given a dataset consisting of high-resolution lidar scans $\mathcal{P}_{HR}$. We downsample $\mathcal{P}_{HR}$ to a lower resolution by skipping a number of lines, and we train different models to reconstruct $\mathcal{P}_{HR}$. We use the range image representation for the downsampling operation since each row in $\mathcal{Q}$ represents a lidar beam. Moreover, it is an invertible transformation, meaning $\mathcal{P}_{LR}$ can be obtained by uniformly downsampling the rows of $\mathcal{Q}_{HR}$ and projecting $\mathcal{Q}_{LR}$ back to 3D. 

\noindent\textbf{Experimental Setup}. We train all models on a $\times 4$-upsampling factor in a self-supervised way. To downsample the high-resolution ground truth by a factor of 4, every $4\ts{th}$ row in $\mathcal{Q}_{HR}$ is sampled, following the convention in lidar upsampling works~\cite{lidarsrresnet, unetlidar, iln}. While grid-based methods~\cite{lidarsrresnet, unetlidar, iln} upsample $\mathcal{Q}_{LR}$ into $\hat{\mathcal{Q}}_{HR}$ then compute $\hat{\mathcal{P}}_{HR}$ using the inverse of \cref{eq:rangeimage}, point-based~\cite{punet, pugan, argcn, patchupsample} methods directly upsample $\mathcal{P}_{LR}$ into $\hat{\mathcal{P}}_{HR}$. 

\noindent\textbf{Baselines} We train and evaluate point-based methods~\cite{lidarsrresnet, unetlidar, iln} and grid-based methods~\cite{lidarsrresnet, unetlidar, iln}. All models are trained with their original hyperparameters using their officially published codes. Note that the models \cite{uqlidar, swd, sga} are not included as the code was not available to run the comparison. In our benchmark, we also include SWIN-IR~\cite{swinir}, a state-of-the-art transformer-based image super-resolution model. SWIN-IR is trained on range images similar to the other grid-based models.

\noindent\textbf{Dataset} We choose the Kitti Raw dataset \cite{kitti} as the benchmark. The lidar sensor in Kitti is the Velodyne HDL 64-E, which has a 64-lines resolution and around $2048$ points approximately per line. However, processing this resolution with point-based methods is computationally hefty and a model with batchsize of $1$ does not even fit into the used GeForce RTX 2080 Ti GPU. To ensure a fair comparison between all algorithms, we limit the size of the pointcloud to $10k$ points ($40 \times 256$) using the preprocessing of \cite{deepgm}. We use the train/validation/test $(40k/80/700)$ split used by \cite{deepgm}. 

\input{Figures/stats}
\noindent\textbf{Metrics} While grid-based methods for lidar upsampling have used 2D-metrics like mean absolute error (MAE) or root mean square error (RMSE) on the range image to evaluate the performance, point-based methods have mainly used the earth-moving distance (EMD) and chamfer distance (CD) \cite{pointsetgeneration}, because they measure distance between subsets in $\mathcal{R}^{3}$.  Moreover, it has been shown in \cite{learningrepresentations} that EMD strongly correlates with perceptual quality. For this reason, we choose EMD and CD as the unified metrics to benchmark all methods on the Kitti Raw dataset. We present the results in \cref{table:kitti}.

\noindent\textbf{Results of the comparative study.} From \cref{table:kitti}, we can observe that point-based methods underperform on lidar upsampling, especially when trained from scratch on lidar. All evaluated point-based models, except PUNet, do not converge during training. For further investigation, we evaluate point-based models pretrained on a dataset of synthetic single object pointclouds proposed by PUNet~\cite{punet}. Surprisingly, PUGAN and ARGCN perform better when pretrained on this dataset than when trained on the pointclouds directly. We highlight that during pretraining no lidar scan was seen, as the pretraining dataset only contains single object pointclouds. By looking at the upsampled pointclouds in \cref{fig:pointbased}, we can see that, except for PUNet, no point-based method was able to replicate the lidar scan pattern. The upsampled pointclouds exhibit higher density around each line, but no new lines are added in between. We hypothesize that these networks are not able to reproduce the scan pattern by architectural design, and hence do not converge during training. 

\input{Tables/kitti_benchmark}

In order to understand why point-based methods (except PUNet) cannot replicate the scan pattern, we inspect and compare the architectures of point-based models. In general, each network can be modeled as being composed of 3 parts: (i) a feature extraction backbone, (ii) a feature expansion module, and (iii) pointset generation layer. Although these parts are different across all models, it is the feature extractor of PUNet that is distinct from other algorithms due to its hierarchical design. In PUNet, point features are computed at different scales, followed by a multi-level feature aggregation resulting in a wide receptive field that captures both local and global information in the pointcloud. Subsequent architectures like PUGAN, 3PU and ARGCN have small receptive fields, an intentional design choice that outperforms PUNet on single object datasets with a uniform spatial point distribution. However, lidar pointclouds are not uniform by design, and a local receptive field that does not span different vertical lines will lead to the generation of new points located close to the input points. Consequently, the lidar scan pattern cannot be replicated (\cref{fig:pointbased}). Note that there is no straightforward way to enlarge the receptive field of these architectures. For instance, increasing the number of nearest neighbours in the layers of 3PU~\cite{patchupsample} and ARGCN~\cite{argcn} would entail a very high-computational cost, with only a marginal increase in receptive field. Also, there is no guarantee that increasing the number of nearest neighbours for each point will include points from other vertical channels. PUGAN~\cite{pugan} deploys per-point feature extraction layers, which do not have a tunable parameter for the receptive field.

On the other hand, grid-based methods~\cite{lidarsrresnet, unetlidar, iln} benefit from structuring the pointcloud into a range image. This representation has several desirable properties: it is invertible, it provides a 2D dense spatial grid structure to the 3D sparse and unstructured pointclouds, thereby intrinsically modeling the lidar scan pattern, and it can be efficiently processed using the widespread hardware-accelerated convolution operations. Their good performance can be attributed to the fact that even convolutional filters with small receptive fields span different lidar beams vertically. Results in \cref{table:kitti} show little difference between all 4 models.  

This analysis motivates us to adopt a grid-based approach as well. However, grid-based models have weaknesses: they suffer from the smoothing effects of convolutional layers, which blur edges and sharp object boundaries~\cite{unetlidar}. These effects are mitigated to some extent in the previous works either by leveraging pretrained point segmentation networks~\cite{lidarsrresnet} (which incurs extra label costs), by filtering uncertain points~\cite{unetlidar, uqlidar} or by using an interpolation approach~\cite{iln}. We take a different approach by first analyzing the lidar scan geometry and point distribution.

\input{Figures/method}
\section{Proposed Methodology}
\label{sec:method}
In this section, we first take a closer look at the range image representation. Then, we introduce our framework, HALS, Height-Aware Lidar Super-resolution. We present the generator's architecture, the used range image representation and loss functions.

\subsection{The Range Distribution of Different Beams}
\label{sec:stats}
It is well-established in image super-resolution~\cite{srgan}, that minimizing the $\mathcal{L}_1$-loss with a CNN architecture leads to blurring artifacts and neglecting high-frequencies from edges and corners. A close visual inspection of a range image reveals even more challenges compared to camera images. First, range images contain black lines or streaks, which correspond to no-return from the laser sensors. The juxtaposition of empty regions and foreground objects create high frequencies in the 2D image space. CNNs might find it hard to process the empty regions, leading to the generation of noisy points. Second, far away objects appear to be constrained or squished in the top rows, suggesting that the upper and lower parts of the range image have different range distributions. 

To confirm this second observation, we measure the average range value and standard deviation of non-zero points in each row across all examples in a dataset, and report these statistics on 3 datasets in \cref{fig:stats}. The range distribution of the 3 datasets (which come from different lidar sensors in real urban scenes) show a similar tendency. The upper rows of the image have on average a higher mean range and standard deviation than the middle and bottom rows. This means that the upper region has a wider distribution of range values, and consequently higher spatial frequencies.

\subsection{Height-Aware Generator Architecture}
Our goal is to design a generator architecture that is sensitive to the height of the range image. We hypothesize that the upper and lower part should be upsampled differently. In the upper part of the range image, an upsampling layer should only observe a small neighborhood around each pixel, while it should be allowed to observe larger neighborhoods in the lower part. The reason for this is that far away objects are compressed in the upper part and appear to have a smaller scale on the 2D image space. Thus, the network should only focus on a small spatial context around each pixel. In contrast, objects in the lower parts are typically closer and bigger. In terms of neural architecture design, the neighborhood observed by the network is called the receptive field. Smaller receptive fields can be obtained by cascading a small number of layers (shallow backbone), while larger receptive fields result from cascading a larger number of layers. By controlling the receptive field size before the upsampling layer, one can decide which points in the range image should be looked at for the super-resolution. 

Our \textit{key idea} is to upsample the full-sized range image multiple times with different receptive fields and fuse the outputs together using confidence maps from each upsampling branch. We find that 2 different receptive fields (a small one and a large one) are enough to cover the range distribution (\cref{sec:ablations}). To this end, we present our architecture in \cref{fig:architecture}. The dimensions of the input range image $\mathcal{Q}_{LR}$ is $[B, C, H, W]$, where $B$ is the batch size, $C$ is the number of features of the range image (defined later) and $H, W$ are the input resolution. The proposed generator has a point encoding layer, a feature extractor which consists of 16 blocks, two upsampling layers and two final regression layers. The point encoding layer is a shared multilayer perceptron (MLP) which transforms the input features to the high dimensional feature space ($64$ channels). The MLP is implemented by a $1 \times 1$ convolutional layer. Contrary to previous works on super-resolution which only have one upsampling layer after the feature extractor, we divide the backbone into 2 smaller connected backbones: a small one with 4 dilated residual blocks (DRBs) and a larger one with 12 DRBs. We add an upsampling layer after each backbone. A regression layer follows each upsampling layer and outputs $C+1$ channels: $C$ channels for the range image features and $1$ channel which features the mask logits. The two masks are then concatenated and passed to a softmax activation function to normalize them, such that the summation of the masks is equal to $1$. 

To summarize, we use two upsampling layers on one shared backbone but with two different receptive fields. Each branch outputs a full-sized range image and a mask. We call the range image and mask from the small backbone $\hat{\mathcal{Q}}_{shallow}$ and $\mathbf{m}_{shallow}$. The output of the other branch consists of $\hat{\mathcal{Q}}_{deep}$ and $\mathbf{m}_{deep}$. The final range image is a weighted average of the two branches (\cref{eq:gen}). Note that the masks are not binary: they have continuous values between 0 and 1 to weigh the predictions of each branch.

\begin{equation}
\label{eq:gen}
\begin{aligned}
  & \hat{\mathcal{Q}} = \hat{\mathcal{Q}}_{shallow} \cdot \mathbf{m}_{shallow} + \hat{\mathcal{Q}}_{deep} \cdot \mathbf{m}_{deep} \\  
  & \mathbf{m}_{shallow} + \mathbf{m}_{deep} = 1 \\
\end{aligned}
\end{equation}

One might think that having two generators (one for the upper part and one for the lower part) would lead to the same result. However, having 2 generators would increase the number of parameters which is undesired in real-time applications. Moreover, we show in \cref{sec:ablations} that 2 generators are suboptimal compared to the proposed method. Our height-aware generator namely has two advantages. First, the shallow layers of the backbone learn faster because of the gradient flow from the shallow upsampling branch. This is similar to having an auxiliary loss~\cite{auxloss}. Second, by allowing the upsampling branches to predict the whole range image as opposed to only one part of it, we allow the branches to correct each other. The masks can be interpreted as the upsampling layer's confidence map: low value indicate high epistemic uncertainty and would weight down the branch's prediction. This would not be the case if we have 2 separate generators, as each generator will only learn on some part of the range image, leading to a weaker training signal.    
Finally, to allow for more flexible receptive fields, we choose to replace normal residual blocks inside the backbones by the DRBs, inspired by \cite{rangercnn}. The dilated convolutions allow to capture coarse and fine details at various receptive fields. To make the paper self-contained, we include the block design in \cref{fig:architecture}.

\subsection{Range image representation with polar coordinates}
An assumption often made in the spherical projection (\cref{eq:rangeimage}) is that all lidar beams have equal angular spacing in the vertical direction. This assumption does not always hold, as it depends on the sensor type. The sensor's vertical resolution may vary with the absolute value of the elevation angle. Some sensors exhibit a dense vertical angular resolution in the middle and a sparse resolution in the upper and lower part of the range image. However, the inverse of the spherical projection leads to quantization errors when computing the absolute height of the points, $z = rsin(\theta)$, because it assumes that all elevation angles $\theta$ are regularly spaced. We argue that it is more beneficial for the network to directly observe and learn the polar or cartesian coordinates, similar to the point-based operations, in order to reduce this vertical quantization error. To this end, we represent the lidar scene as a range image with polar coordinates ($\mathcal{Q}(u,v) = (d, z)$), as they empirically show a lower generalization error than cartesian coordinates. This representation is used for the input and output range image features ($C=2$). Note that when projecting the range image back to 3D space, the inverse of \cref{eq:rangeimage} is only applied for $(x,y)$ using the polar range $d$ and assuming equally spaced azimuth angles from $u$, while $z$ value is taken from the upsampling branches and is not estimated from $v$. By directly estimating the $z$ value for each point in the final layer, the generator can better match the real point distribution in the vertical direction. 

\subsection{Surface Normal Loss}
Previous works have mostly used the $\mathcal{L}_1$ loss with respect to the ground truth to train the network. However, the $\mathcal{L}_1$ loss is not sensitive to the high-frequency details in the range image and might lead to noisy object boundaries. To regularize the training, we draw inspiration from monocular depth estimation (MDE), which is the task of estimating a depth map from a camera image. Although MDE is very different from lidar upsampling, we draw a connection between the two fields: both estimate depth values on a 2D plane. In MDE, it has been shown~\cite{surfacenormal} that the $\mathcal{L}_1$ loss does not sufficiently penalize the shift in the estimated edges. To compensate for these shortcomings, some MDE works~\cite{vnl, surfacenormal} estimate the surface normal of the estimated depth map and penalize its deviations from the surface normal of the ground truth. We adopt a robust variant of the surface normal loss, called the virtual normal loss ($\mathcal{L}_{VNL}$) \cite{vnl} in addition to the used $\mathcal{L}_1$ loss. Specifically, $K$ groups of $3$ non-colinear points are sampled from each pointcloud and the normal vector to the plane formed by triplet is computed.
The final loss is a combination of $\mathcal{L}_1$ and $\mathcal{L}_{VNL}$. Let $n_k$ be the normal vector estimated from a group of points in the ground truth, and $\hat{n}_k$ be the normal vector estimated from $\hat{\mathcal{Q}}$ from the same group of points. The final loss used for training can be expressed as:

\begin{equation} 
\label{eq:loss}
\begin{split}
\mathcal{L} &=  \frac{1}{HWC} \sum_{i,j,c}^{H,W,C} \mid\mid \hat{\mathcal{Q}}_{HR}(u,v,c) - \mathcal{Q}_{HR}(u,v,c) \mid\mid_1 \\
& +\frac{1}{K}\sum_{k=1}^K \mid\mid \hat{n}_k - n_k \mid \mid_1 
\end{split}
\end{equation}
Note that we use the $\mathcal{L}_1$ loss on the polar coordinates $(d, z)$. We summarize the whole training pipeline in \cref{fig:pipeline}

\section{Experiments}
\label{sec:results}
In this section, we demonstrate that the proposed approach achieves improved generation quality on 3 real-world autonomous driving datasets. We perform extensive ablation studies on 2 datasets. Furthermore, we show the performance of an object detection model on upsampled lidar scans generated from the proposed model.

\subsection{Experimental Setup}
We conduct experiments on 3 datasets: Kitti Raw which was previously introduced, Kitti Object~\cite{kitti} and Nuscenes~\cite{nuscenes}. The lidar sensor used in Kitti Object and Kitti Raw dataset is the Velodyne HDL-64E which has 64 beams and a vertical field of view of $26.8^{\circ}$, while Nuscenes uses Velodyne HDL-32E with 32 beams and a vertical field of view of $40^{\circ}$. AS previously mentioned, the number of points is reduced in Kitti Raw so that the range image size is $40 \times 256$. Larger resolutions are used in Kitti Object and Nuscenes, where the ground truth resolution is $64 \times 700$ for the first and $32 \times 1024$ for the second. We choose these resolutions to be able to calculate the EMD metric, as larger pointcloud sizes do not fit into the used GPU (GeForce RTX 2080 Ti). We use a train/validation/test of $(3612/100/3769)$ on Kitti Object, $(40k/80/700)$ on Kitti Raw and $(27k/1k/6k)$ on Nuscenes. A batchsize of $32$ is used for Kitti Raw, while a batchsize of $24$ is used for Kitti Object and Nuscenes. We use the ADAM optimizer with an initial learning rate of $0.0001$ that is decayed by a factor of 0.5 every 40 epochs on Nuscenes and Kitti Raw and every 80 epochs on Kitti Object. 

\input{Figures/pipeline}
\input{Tables/ablation_kraw}
\input{Tables/ablation_generator_design}
\input{Figures/masks}
To evaluate the upsampling quality, we use the EMD and CD. Moreover, we add 6 metrics that were used in previous works. We calculate the MAE and RMSE between the generated and ground truth range images. MAE and RMSE were used in previous works on grid-based lidar upsampling~\cite{lidarsrresnet, unetlidar, iln, uqlidar}. Similar to ILN~\cite{iln}, generated and ground truth pointclouds are voxelized (using a voxel size of $0.1 m \times 0.1 m \times 0.1 m$). If one or more points fall inside a voxel, it is assigned a value of $1$, else it is assigned $0$. Then, we calculate the Intersection-over-Union (IoU), the Precision, Recall and F1-score with respect to the ground truth. These 4 metrics measure the 3D alignment of the generated pointcloud with the ground truth in a \textit{coarser} way than EMD and CD (which measure the difference between point distributions). They indicate how much the structure of the pointcloud is similar to the high-resolution real pointcloud.

\input{Tables/mainresults}
\subsection{Ablation studies}
\label{sec:ablations}
To showcase the importance of our contributions, we ablate different parts of our model. We make an incremental component analysis on Kitti Raw dataset ($4\times$ upsampling rate) in \cref{tab:ablation_kraw} and an ablation on the generator design on the Nuscenes dataset ($2\times$ upsampling rate) in \cref{tab:ablation_generator_design}. We also show a visualization of the generated masks ($\mathbf{m}_{shallow}$ and $\mathbf{m}_{deep}$) in \cref{fig:masks}.

\noindent\textbf{Incremental Component Analysis} We build our framework on top of SRResNet~\cite{srgan}, a widely used image super-resolution architecture.  We incrementally show the effects of the proposed contributions on the Kitti Raw dataset ($4 \times$ upsampling rate). In configuration 1 in \cref{tab:ablation_kraw}, we start with a straightforward application of SRResNet on range image with spherical coordinates and $L_1$ loss only. In configuration 2 and 3, we change the input and output coordinates to cartesian and polar respectively. We already notice a considerable decrease in all 3D metrics. Note that polar coordinates show a higher empirical performance than cartesian coordinates. We hypothesize this happens because it is easier for the network to regress 2 variables $(d,z)$ than 3 $(x,y,z)$. In configuration 4, we add $\mathcal{L}_{VNL}$ and notice a decrease in MAE and an increase in IoU, Precision, Recall and F1-score. Replacing residual blocks with DRBs slightly improves the EMD and CD but the other metrics drop. Finally, we replace the original generator with the proposed height-aware generator, which shows a considerable improvement in the last 4 metrics and an equal or sightly superior performance on the first 4 metrics.

\noindent\textbf{Ablation on the Generator Design.} In \cref{tab:ablation_generator_design}, we study several design choices in the generator design on the Nuscenes Dataset ($2 \times$ upsampling rate). Note that the Nuscenes dataset is harder than Kitti Raw and Kitti Object, as it has fewer lines, more sparse regions in the range image and a wider vertical field of view. We start with our baseline with polar coordinates, DRB, and $\mathcal{L}_{VNL}$ in configuration A. We refer to this model as generator (A). In configuration B, we use 2 generators (A), one on the upper part of the range image (from row $0$ to $3$) and one on the lower part only (from row $4$ to $15$). Both generators have 16 DRBs. We notice an improvement in most metrics, confirming our hypothesis that different parts of the range image need different upsampling models as they exhibit different properties. In configuration C, we reduce the receptive field of the generator for the upper part (4 DRBs only), but we notice a slight decrease in 3D metrics compared to configuration B. This could be attributed to the low model capacity of the upper generator. Then in configurations D, E and F, we try different settings for our height-aware generator. Namely, in configuration D, we place 3 upsampling branches instead of 2, after blocks number 4, 8, and 16. In configuration E, we place 2 branches at blocks 12 and 16, while in configuration F, they are placed at blocks 8 and 16. Finally, we show the proposed model, which features 2 branches at blocks 4 and 16 in configuration G. Specifically, HALS has a superior performance compared to 2 generators in configurations B and C, while having fewer parameters. Moreover, placing the first upsampling branch after a small number of blocks shows superior performance than placing it after a larger number of blocks (configurations E and F), highlighting the importance of the receptive field as a design parameter. 

\noindent\textbf{What are the masks focusing on?} We hypothesized that a lower receptive field for the first upsampling branch smaller be beneficial for the higher part of the range image. To confirm the soundness of our hypothesis, we visualize the masks $\mathbf{m}_{shallow}$ and $\mathbf{m}_{deep}$ of a scene in the Kitti Object dataset in \cref{fig:masks}. The mask from the shallow branch has higher values (yellow and orange in the heatmap) than the mask from the deep branch in the upper part. This implies more weight is given to the prediction from the shallow branch than the deep branch in this part, as it is more confident in its generated range image. In the lower part of the range image, $\mathbf{m}_{deep}$ has more contribution than $\mathbf{m}_{shallow}$. 

\input{Figures/cars}

\subsection{Main Results}
\label{sec:main_results}
We compare the proposed approach with the state-of-the-art grid-based models in \cref{tab:main} on 3 datasets, with different resolutions and upsampling rates. We include all grid-based baselines and we add to them the SWIN-IR~\cite{swinir}, as it is a state-of-the-art image super-resolution model. HALS outperforms the baselines on the majority of metrics, sometimes by a significant margin (especially in EMD, CD and IoU). Note that 2D metrics are less important than 3D, since the pointcloud lies in the 3D space. For instance, SWIN-IR has the lowest RMSE and MAE on Kitti Object and Kitti Raw but this does not translate to the best performance in 3D metrics. On Nuscenes, ILN has a good performance on IoU, Precision, Recall and F1-score but bad performance on EMD and CD. On the other hand, SWIN-IR is better on EMD and CD but lacks behind in IoU and F1-score. The proposed model achieves strong results on all 6 3D metrics simultaneously. In \cref{fig:qualitative_scenes}, we show qualitative results for upsampled pointclouds from 3 different baselines and our model on the Kitti Object dataset ($\times 4$ upsampling rate). For illustration purposes, we show the front part of the pointcloud. We observe that LIDAR-SR and SWIN-IR generate pointclouds with noisy shapes. Pointclouds from ILN exhibit clusters of high density and other clusters with low density, showing an overall point distribution different from the ground truth. It can also be seen that some lines are generated very close to the existing input lines. Since ILN is an interpolation approach that generates new points using a weighted average of the coordinates of their nearest neighbours, it can become susceptible to artifacts caused by the height-dependent range distribution. In contrast, generated pointclouds from our method have a similar point distribution as the ground truth and objects with more plausible shapes. 

\subsection{Object Detection Results}
To measure how well the shape of foreground objects is preserved during upsampling, we evaluate the performance of an object detection model on the upsampled pointclouds. Specifically, we train Pointpillars~\cite{pointpillars} to detect cars in high-resolution pointclouds from the Kitti Object dataset~\cite{kitti}. Then, we evaluate the performance of the model on upsampled pointclouds from 3 models: LIDAR-SR~\cite{unetlidar}, ILN~\cite{iln} and the proposed HALS model. We use the official Kitti evaluation protocol and report the Average Precision (AP) with 40 recall positions at an overlap threshold of $0.7$ IoU. The proposed model outperforms both ILN and Lidar-SR on the Easy, Moderate and Hard categories. In Figure~\ref{fig:cars}, we extract cars from the upsampled pointclouds using their ground truth bounding boxes and visualize them. The car from LIDAR-SR is noisy and has few points in the upper part. ILN and SWIN-IR generate more points but their shape is different from the ground truth. For instance, ILN repeats the same line in the lower part of the car, generating a cuboid-like shape. In contrast, HALS is able to better approximate the shape of the car in the dataset.

\input{Tables/odmetrics}
\input{Figures/qualitative.tex}

\subsection{Discussion and Limits}
The results of the benchmark that we presented in \cref{sec:overview} are comprehensive, but they are not conclusive. Although current grid-based methods outperform point-based approaches, future works can focus on improving point-based methods. However, the biggest challenge would be how to make point-based methods suitable for real-time deployment as they require more memory and computation time. Scaling to a large number of points is still a challenge. 

Lidar upsampling can be used to increase the performance of many downstream computer vision applications (object detection, point segmentation...) all while using low-resolution sensors. In this work, we do not explore these applications, as it is a topic that deserves to be addressed on its own. Therefore, we leave it to future works.  

Another limitation of the proposed approach is that it learns a deterministic mapping: the generator can synthesize only one high-resolution pointcloud from the low-resolution input. Future works can focus on how to apply generative models like generative adversarial networks, flows or diffusion models to lidar upsampling in order to learn a conditional probability distribution. 

\section{Conclusion}
\label{sec:Conclusion}
In this work, we have benchmarked the performance of different point-based and grid-based methods in the lidar upsampling tasks on the Kitti Raw dataset. Our analysis revealed the superiority of grid-based methods due to their vertical receptive field which spans different beams. We have also shed light on a peculiar characteristic of the range images; namely, the range distribution varies with the beam height, starting with a high mean and standard deviation at the top, which gradually decrease towards the bottom. We have proposed a generator architecture to match this height-dependent range distribution. By assigning varying receptive fields to different vertical parts of the range image, the generator collects the necessary spatial information to upsample the scene while preserving its shape. We have also changed the network's input and output representation to polar coordinates to explicitly generate the points height information. Finally, we have adopted a surface normal loss to preserve the 3D structure. With the proposed contributions, HALS sets a new standard for lidar upsampling on 3 real-world challenging datasets. Extensive ablation studies were conducted on 2 datasets to validate the soundness of our design choices. 

\clearpage
\clearpage

\begingroup
\setstretch{0.9}
\setlength\bibitemsep{0pt}
\printbibliography
\endgroup


\end{document}

%% file: Figures/teaser.tex
\begin{figure*}
    \centering
    \includegraphics[width=0.90\textwidth]{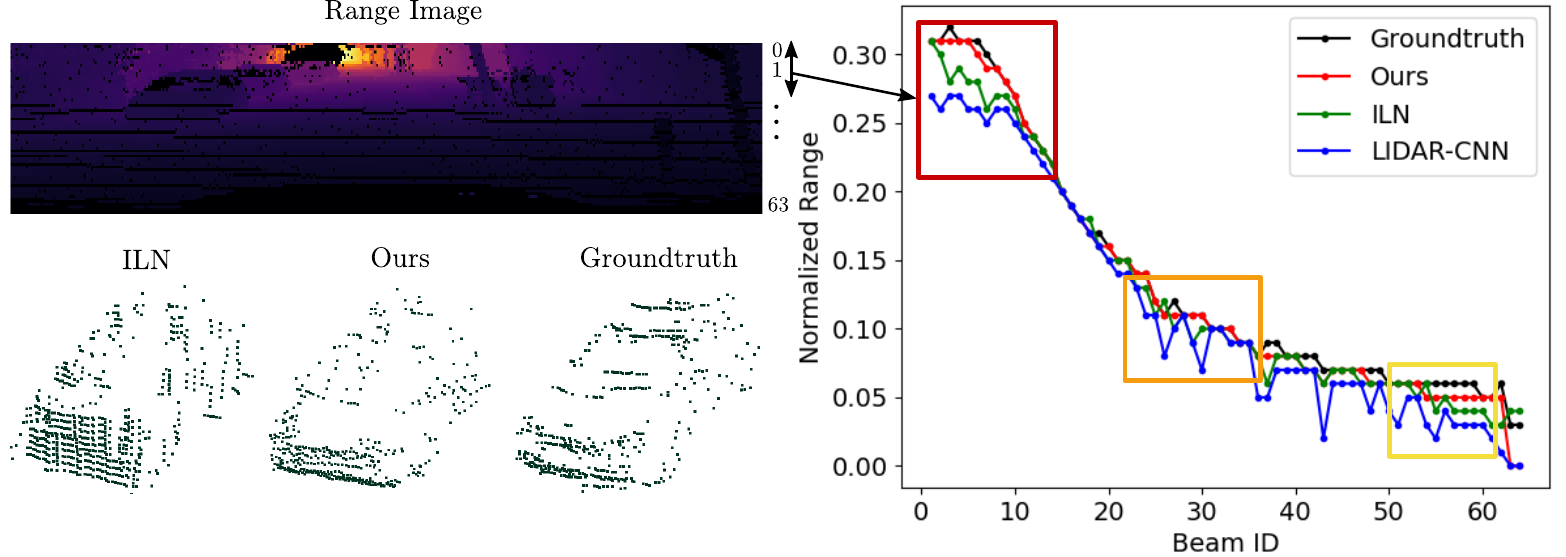}
    \caption{When projected on a 2D spherical range image, a lidar scan exhibits a height-dependent range distribution. We find that far away objects (high range values) are usually represented in the upper part of the range image (Beam ID 0 corresponds to the highest row in the range image). Upsampled lidar scans should also follow this distribution. We record the average range per beam ID of generated pointclouds from our height-aware generator and $2$ state-of-the-art upsampling models on the Kitti Object Dataset and show that we can better follow the ground truth distribution. Extracted cars from upsampled lidar scans demonstrate that the overall geometry and shape of foreground objects are better preserved. }
    \label{fig:teaser}
    \vspace{-1em}
\end{figure*}

%% file: Figures/rw.tex
\begin{table}
\resizebox{\columnwidth}{!}{\begin{tabular}{| l | c | c |} 
\hline
& Point-based methods & Grid-based methods \\
\hline
Pointcloud Upsampling & \cite{punet, pugan, patchupsample, argcn} & NA \\
\hline
Lidar Upsampling & \cite{punet, pugan, patchupsample, argcn} \cite{swd, sga} & \cite{lidarsrresnet, unetlidar, iln, uqlidar} \\
\hline
\end{tabular}}
\caption{Overview of the discussed related works. Lidar upsampling is a sub-task of pointcloud upsampling. While both can use point-based methods, grid-based methods which use a spherical projection are only applicable for lidar upsampling. We evaluate both methods for lidar upsampling.}
\label{fig:rw}
\vspace{-1em}
\end{table}

%% file: Figures/results.tex
\begin{figure*}[t]
    \centering
    \captionsetup[subfloat]{position=top, labelformat=empty, skip=0pt}

    \begin{minipage}{0.90\textwidth}
    \subfloat[Input]{\includegraphics[width=  0.20\textwidth, height=  0.2\textwidth]{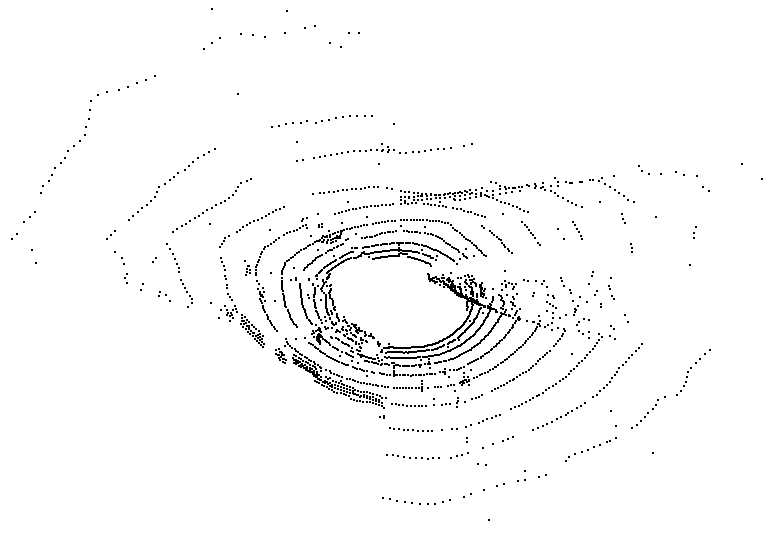}} \hfill
    \subfloat[ARGCN \cite{argcn}]{\includegraphics[width=  0.20\textwidth, height=  0.2\textwidth]{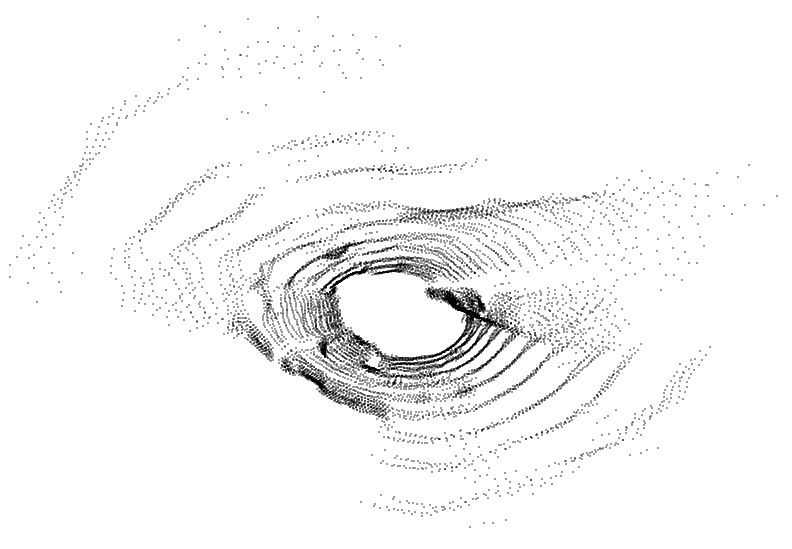}} \hfill
    \subfloat[PUGAN \cite{pugan}]{\includegraphics[width=  0.20\textwidth, height=  0.2\textwidth]{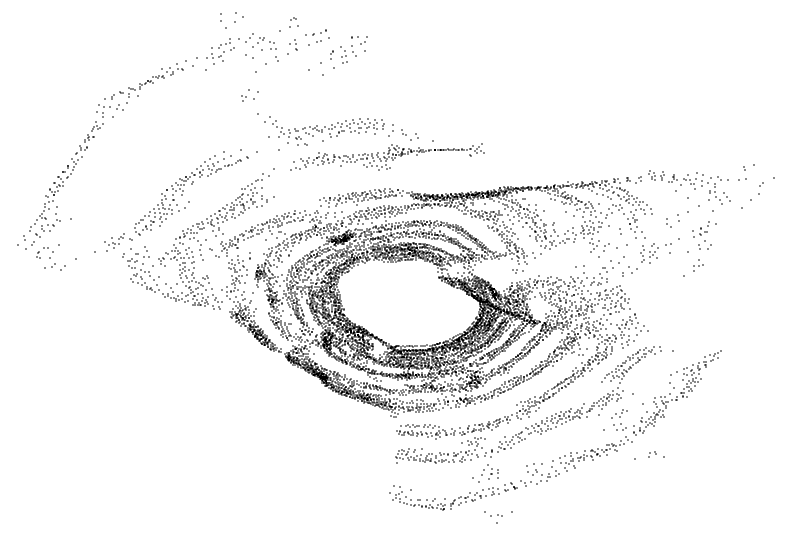}} \hfill
    \subfloat[PUNet \cite{punet}]{\includegraphics[width=  0.20\textwidth, height=  0.2\textwidth]{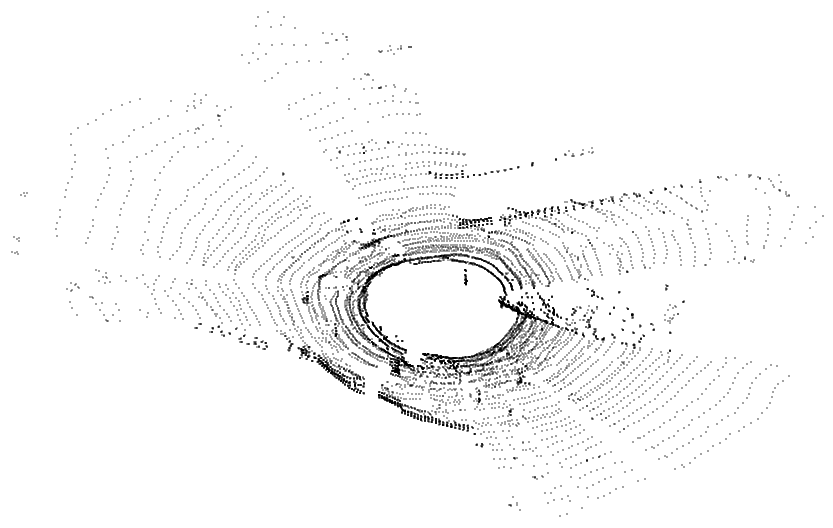}} \hfill
    \subfloat[Groundtruth]{\includegraphics[width=  0.20\textwidth, height=  0.2\textwidth]{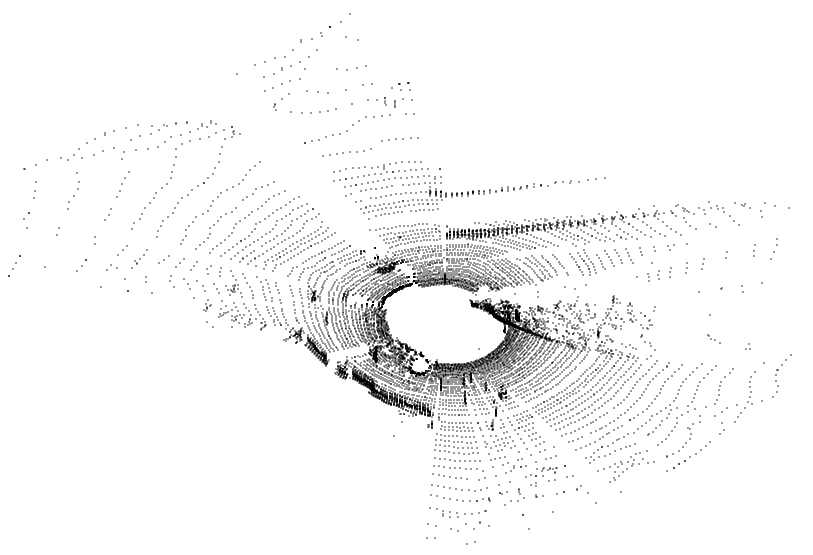}} \vfill
    \end{minipage}
    \caption{Generated lidar scans from point-based models. Only PUNet is able to replicate the lidar scan pattern. }
    \label{fig:pointbased}
    \vspace{-1em}
\end{figure*}


%% file: Figures/stats.tex
\begin{figure*}
\centering
\includegraphics[width=0.95\textwidth]{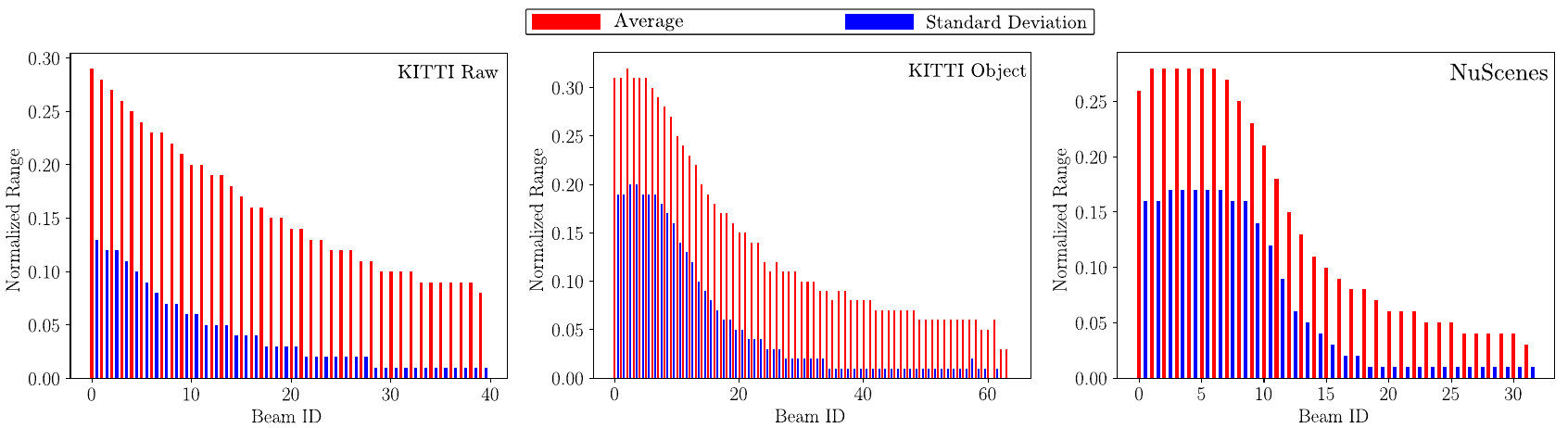}
\caption{We record the average range and standard deviation per beam (a beam corresponds to a row in the range image.) Beam 0 represents the highest one from the ground. We note that the range distribution exhibits a height-dependent behaviour: far away objects are mostly represented in the upper part of the range image. Also, the standard deviation of range is bigger in the upper rows, suggesting they are rich in high spatial frequencies.}
\label{fig:stats}
\end{figure*}

%% file: Tables/kitti_benchmark.tex
\begin{table}[t!]
\small
\begin{threeparttable}
    \centering 
    \begin{tabular}{c l | c | c c} 
    \Xhline{1pt}
    \multicolumn{2}{c|}{\textbf{Framework}} & \textbf{Pretrained\tnote{1}} & \textbf{EMD} $\downarrow$ & \textbf{CD} $\downarrow$  \\ 
    \hline
    & 3PU \cite{patchupsample}                    & \cmark  & 1265  & 6.58   \\
    & 3PU \cite{patchupsample}                    & \xmark  & 924   & 6.37   \\
    & PUGAN \cite{pugan}                          & \xmark  & 866   & 77.82  \\
    Point-  & ARGCN\cite{argcn}                   & \xmark  & 829   & 29.07  \\
    based   & PUGAN \cite{pugan}                  & \cmark  & 385   & 0.86   \\
    & PUNet \cite{punet}                          & \cmark  & 371   & 1.70   \\
    & ARGCN \cite{argcn}                          & \cmark  & 265   & 0.73   \\
    & PUNet \cite{punet}                          & \xmark  & 241.6 & 0.67   \\
     \hline
    Grid- & LIDAR-CNN \cite{lidarsrresnet}        & \xmark  & \textbf{100.1}   &  0.052 \\
    based & LIDAR-SR \cite{unetlidar}                      & \xmark  & 101.0 & 0.054  \\
    & ILN~\cite{iln} & \xmark & 104.2 & 0.061 \\
    & SWIN~\cite{swinir} & \xmark & 101.0 & \textbf{0.051} \\
    \Xhline{1pt}
    \end{tabular}
    \begin{tablenotes}
    \item[1] Pretrained on a synthetic dataset
    \end{tablenotes}   
    \caption{Quantitative comparisons on Kitti Benchmark.}
    \label{table:kitti}
    \vspace{-1em}
\end{threeparttable}
\vspace{-1em}
\end{table}

%% file: Figures/method.tex
\begin{figure*}
    \centering
    \includegraphics[width=0.98\textwidth]{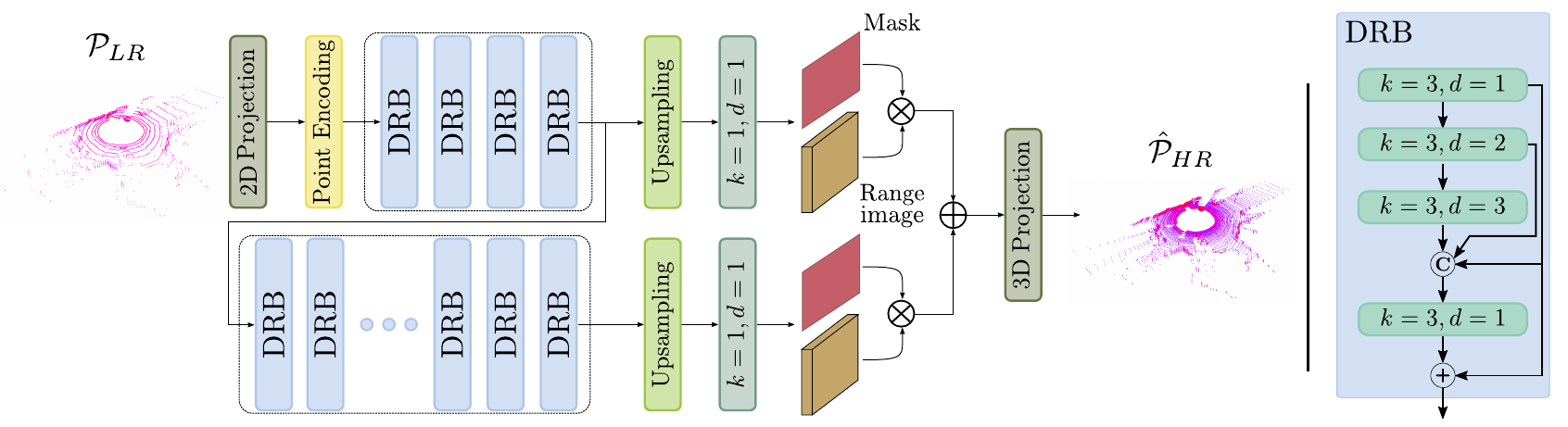}
    \caption{\textit{Left}: The proposed generators architecture. We upsample the pointcloud at two locations in the backbone, intentionally chosen to have a local and a global receptive field. Both outputs are fused using confidence maps from each branch. \textit{Right}: The architecture of the DRB used in the backbone to provide flexible receptive fields.}
    \label{fig:architecture}
    \vspace{-1em}
\end{figure*}

%% file: Figures/pipeline.tex
\begin{figure}
    \centering
    \includegraphics[width=0.95\linewidth]{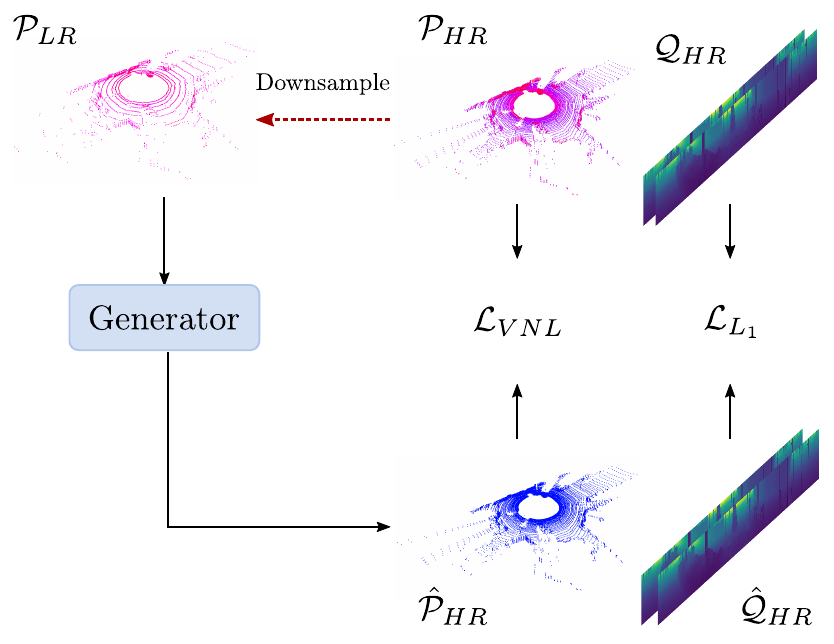}
    \caption{The training pipeline of the proposed model. The generator is trained with an $\mathcal{L}_1$ loss on the range images with polar coordinates and $\mathcal{L}_{VNL}$ on the pointclouds to preserve the structure in 3D.}
    \label{fig:pipeline}
    \vspace{-1em}
\end{figure}

%% file: Tables/ablation_kraw.tex
\begin{table*}[t]
\centering
\renewcommand \arraystretch{1.4}
\scalebox{0.85}{\begin{tabular}{l | c c c c c c c c}
\hline
Model & EMD $\downarrow$ & CD $\downarrow$ & MAE $\downarrow$ & RMSE $\downarrow$ & IoU $\uparrow$ & Precision $\uparrow$ & Recall $\uparrow$ & F1-score $\uparrow$\\
\hline

1- Baseline~\cite{srgan} & 101	& 0.052 & 0.186 & 0.86 & 0.393 & 0.564 & 0.564 & 0.564 \\
\hline
2- Baseline + cartesian & 90.2  & 0.031 & 0.224 & 0.85 & 0.268 & 0.421 & 0.423 & 0.422 \\
3- Baseline + polar     & 84.7  & 0.020 & 0.186 & \textbf{0.84} & 0.443 & 0.614 & 0.614 & 0.614 \\
\hline
4- Baseline + polar + $\mathcal{L}_{VNL}$  & 82.5  & 0.018 & 0.172 & \textbf{0.84} & 0.478 & 0.645 & 0.648 & 0.647 \\
5- Baseline w/ DRB + polar + $\mathcal{L}_{VNL}$  & \textbf{81.4}  & 0.016 & 0.174 & 0.88 & 0.436 &	0.608 &	0.606 &	0.607 \\

6- Height-aware generator + polar + $\mathcal{L}_{VNL}$  & 82.0  & \textbf{0.015} & \textbf{0.171} & 0.88 & \textbf{0.510} & \textbf{0.672} & \textbf{0.671} & \textbf{0.671} \\

\hline
\end{tabular}}
\caption{Ablation study of our model performed on the KITTI Raw Dataset, with $\times 4$ upsampling rate (output resolution = $40 \times 256$).}
\label{tab:ablation_kraw}
\end{table*}

%% file: Tables/ablation_generator_design.tex
\begin{table*}[t]
\centering
\renewcommand \arraystretch{1.4}
\scalebox{0.85}{\begin{tabular}{l l | c c c c c c c c}
\hline
Config. & Model & EMD $\downarrow$ & CD $\downarrow$ & MAE $\downarrow$ & RMSE $\downarrow$ & IoU $\uparrow$ & Precision $\uparrow$ & Recall $\uparrow$ & F1-score $\uparrow$\\
\hline
A & Baseline~\cite{srgan} w/ DRB + polar + $\mathcal{L}_{VNL}$ & 357 &	0.196 &	0.738 &	4.68 &	0.457 &	 0.614 & 0.639 &	0.626 \\
\hline 
B & 2 generators (A) same receptive field & 349 &	0.179 &	0.71 &	4.81 & 0.491 & 0.648 & 0.666 & 0.657 \\
C & 2 generators (A) diff. receptive field& 368 &	0.189 &	\textbf{0.69} &	\textbf{4.67} & 0.480 & 0.639 & 0.665 & 0.652 \\
\hline
D & (A) w/ 3 upsampling layers at (4,8,16) & 344 &	0.180 &	\textbf{0.69} &	4.70 & 0.487 & 0.644 & 0.664 & 0.654 \\
E & (A) w/ 2 upsampling layers at (12, 16) & 351 &	0.186 &	0.71 &	4.73 & 0.471 & 0.626 & 0.654 & 0.640 \\
F & (A) w/ 2 upsampling layers at (8, 16) & 343 &	0.178 &	\textbf{0.69} &	4.70 & 0.493 & 0.653 & 0.668 & 0.661 \\
\hline
G & (A) w/ 2 upsampling layers at (4, 16) - Ours & \textbf{338} &	\textbf{0.171} &	\textbf{0.69} &	4.72 & \textbf{0.505} & \textbf{0.664} & \textbf{0.676} & \textbf{0.670} \\

\hline
\end{tabular}}
\caption{Ablation study on the generator design performed on the Nuscenes Dataset, with x2 upsampling rate (output resolution = $32 \times 1024$).}
\label{tab:ablation_generator_design}
\vspace{-1em}
\end{table*}

%% file: Figures/masks.tex
\begin{figure*}
    \centering
    \includegraphics[width=1.0\textwidth]{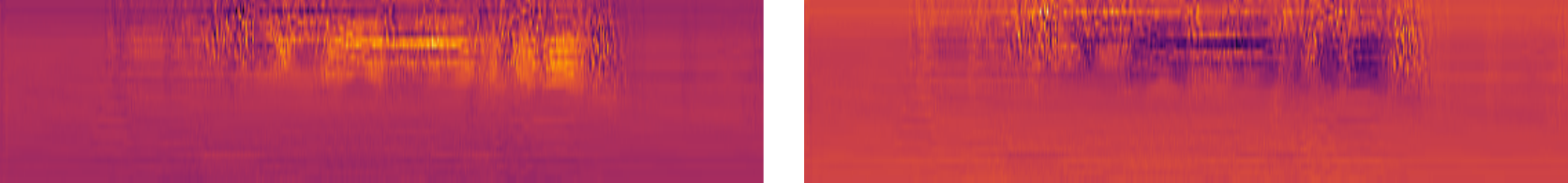}
    \caption{Visualization of the masks produced by the HALS generator (Nuscenes dataset). \textit{Left}: mask from shallow branch. \textit{Right}: mask from the deep branch. Brighter colors indicate higher values (close to 1).}
    \label{fig:masks}
    \vspace{-1em}
\end{figure*}

%% file: Tables/mainresults.tex
\begin{table*}[t]
\centering
\renewcommand \arraystretch{1.4}
\scalebox{0.85}{\begin{tabular}{l c c c c c c c c}
\hline
Model & EMD $\downarrow$ & CD $\downarrow$ & MAE $\downarrow$ & RMSE $\downarrow$ & IOU $\uparrow$ & Precision $\uparrow$ & Recall $\uparrow$ & F1-score $\uparrow$\\
\hline
\multicolumn{9}{c}{KITTI Raw Dataset 4x Output Resolution: $40 \times 256$} \\
\hline

Bilinear & 173 & 0.110 & 0.62 &	1.30 & 0.097 & 0.177 &	0.174 &	0.176\\
LIDAR-CNN~\cite{lidarsrresnet} & 101 & 0.052 & 0.19 &	0.86 & 0.393 & 0.564 & 0.564 & 0.564\\
LIDAR-SR~\cite{unetlidar} & 130	& 0.162 & 0.39 &	2.03 &	0.342 &	0.515 &	0.506 &	0.51\\
SWIN-IR~\cite{swinir} & 101 &	0.051 &	0.19 &	\textbf{0.85} &	0.451 &	0.621 &	0.621 &	0.621\\
ILN~\cite{iln} & 104 & 0.061 &	0.23 &	0.93 &	0.392 &	0.588 &	0.54 & 0.563\\
Ours & \textbf{82} &	\textbf{0.015}  & \textbf{0.17}  & 0.89 &	\textbf{0.510} & \textbf{0.672} & \textbf{0.671} & \textbf{0.671}\\
\hline
\multicolumn{9}{c}{KITTI Object Dataset 4x Output Resolution: $64 \times 700$} \\
\hline
Bilinear & 834 & 0.270 & 1.30 & 3.94 & 0.110 & 0.181 & 0.203  & 0.191\\
LIDAR-CNN~\cite{lidarsrresnet}   & 390 & 0.105 & 0.48 & 2.98 & 0.256 & 0.400 & 0.410 & 0.411\\

LIDAR-SR~\cite{unetlidar} & 757 &	0.110 &	1.113 &	5.59 &	0.277 &	0.447 & 0.419 &	0.432\\
SWIN-IR~\cite{swinir} & 391 & 0.105 & \textbf{0.44} & \textbf{2.81} &	0.376 &	0.537 &	0.554 &	0.545 \\
ILN~\cite{iln} & 629 & 0.101 & 0.49 & 2.97 & 0.336 &	0.501 &	0.504 &	0.502\\
Ours & \textbf{369} & \textbf{0.09} & 0.45 & 3.01 &	\textbf{0.402} & \textbf{0.567} & \textbf{0.573} &	\textbf{0.57} \\
\hline
\multicolumn{9}{c}{Nuscenes Dataset 2x Output Resolution: $32 \times 1024$} \\
\hline
Bilinear & 595 & 0.89 & 1.53 & 5.31 & 0.106  & 0.181 & 0.201 & 0.19  \\
LIDAR-CNN~\cite{lidarsrresnet} & 388 & 0.231 & 0.82  & 4.97 & 0.317   & 0.467 & 0.493 & 0.48\\

LIDAR-SR~\cite{unetlidar} & 514 & 0.209 & 1.39 & 6.77 & 0.200 & 0.340 & 0.325   & 0.332 \\

SWIN-IR~\cite{swinir} & 373 & 0.210 & 0.75 & 4.90 & 0.332 & 0.489 & 0.506   & 0.498 \\

ILN~\cite{iln} & 383 & 0.198 & 0.73 & 4.82 & 0.501 & 0.656 & \textbf{0.678} & 0.667 \\

Ours & \textbf{338} & \textbf{0.171} & \textbf{0.69}  & \textbf{4.72} & \textbf{0.505} & \textbf{0.664} & 0.676 & \textbf{0.670}\\
\hline
\end{tabular}}
\caption{Quantitative comparison against state-of-the art grid-based lidar super-resolution methods.}
\label{tab:main}
\vspace{-1em}
\end{table*}

%% file: Figures/cars.tex
\begin{figure*}
    \captionsetup[subfloat]{position=top, labelformat=empty, skip=0pt}
    
    \centering
    \begin{minipage}{.950\textwidth}
    \subfloat[LIDAR-SR\cite{unetlidar}]{\includegraphics[width=  0.20\textwidth, height=   0.15\textwidth]{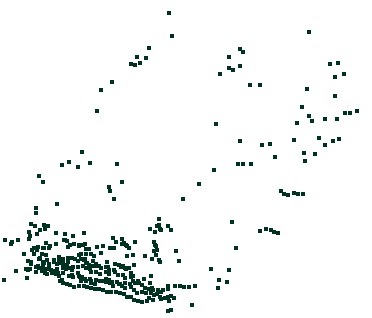}} \hfill
    \subfloat[SWIN-IR\cite{swinir}]{\includegraphics[width=  0.20\textwidth, height=   0.15\textwidth]{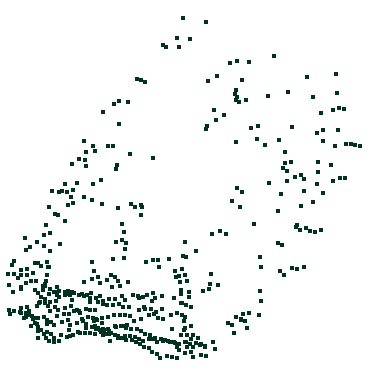}} \hfill
    \subfloat[ILN~\cite{iln}]{\includegraphics[width=  0.20\textwidth, height=   0.15\textwidth]{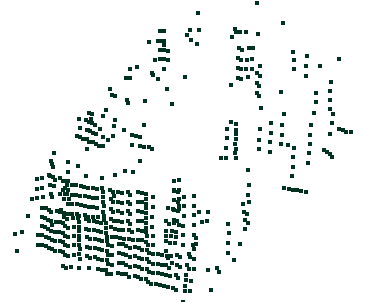}} \hfill
    \subfloat[Ours]{\includegraphics[width=  0.18\textwidth, height=   0.15\textwidth]{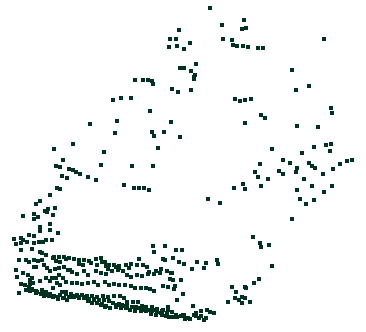}} \hfill
    \subfloat[Groundtruth]{\includegraphics[width=  0.20\textwidth, height=   0.17\textwidth]{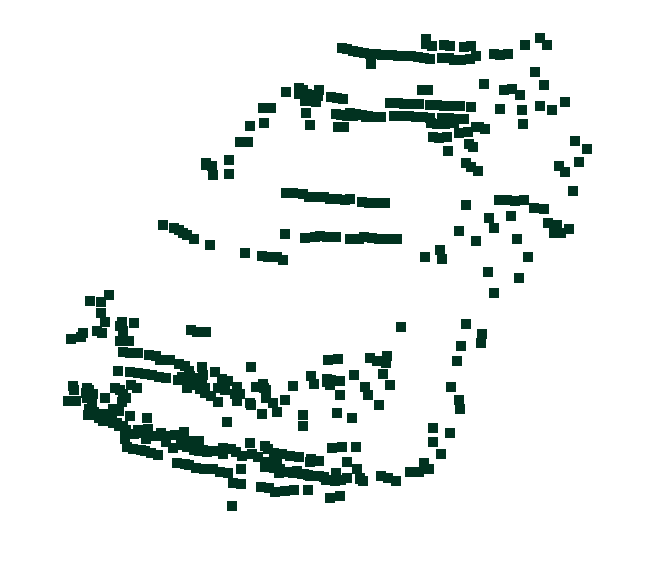}} 
    \end{minipage}

   \caption{Qualitative comparison on Cityscapes dataset}
   \label{fig:cars}
\end{figure*}

%% file: Tables/odmetrics.tex
\begin{table}[t]
\centering
\renewcommand \arraystretch{1.4}
\scalebox{0.85}{\begin{tabular}{l c c c}
\hline
Model & Easy  & Moderate & Hard \\
\hline


LIDAR-SR~\cite{unetlidar} & 44.13	& 25.05 & 20.39 \\


ILN~\cite{iln} & 51.93 & 31.92 &	26.58\\

Ours & \textbf{55.76} &	\textbf{34.00}  & \textbf{27.38} \\
\hline

\hline
\end{tabular}}
\caption{We evaluate a pretrained Pointpillars model~\cite{pointpillars} on $\times4$ upsampled pointclouds from the Kitti Object dataset~\cite{kitti} and report the results on the 'Car' class.}
\label{tab:odmetrics}
\end{table}

%% file: Figures/qualitative.tex
\begin{figure*}[h!]
    \captionsetup[subfloat]{position=top, labelformat=empty, skip=0pt}
    
    \centering
    \begin{minipage}{.80\textwidth}
    \subfloat[Input]{\includegraphics[width=  0.33\textwidth, height=   0.200\textwidth]{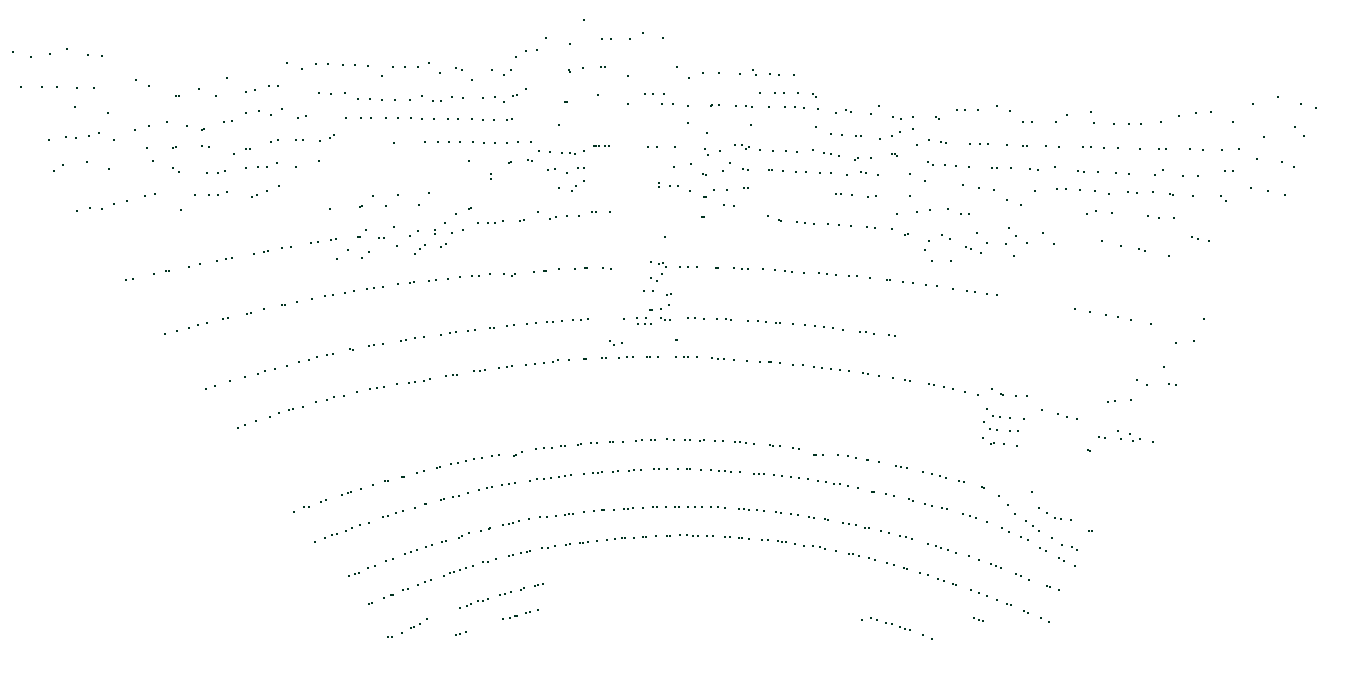}} \hfill
    \subfloat[LIDAR-SR\cite{unetlidar}]{\includegraphics[width=  0.33\textwidth, height=   0.200\textwidth]{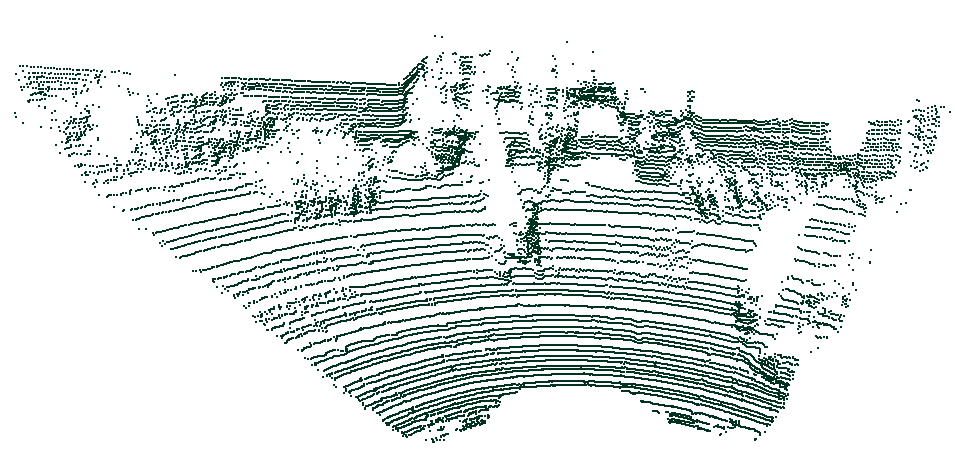}} \hfill
    \subfloat[SWIN-IR\cite{swinir}]{\includegraphics[width=  0.33\textwidth, height=   0.200\textwidth]{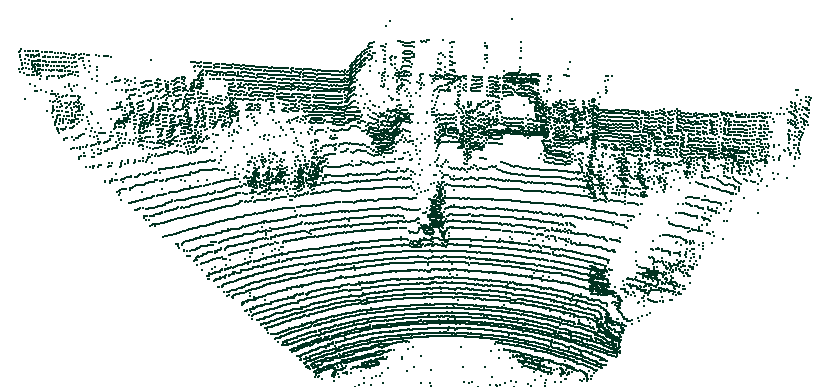}} \vfill
    \subfloat[ILN~\cite{iln}]{\includegraphics[width=  0.33\textwidth, height=   0.200\textwidth]{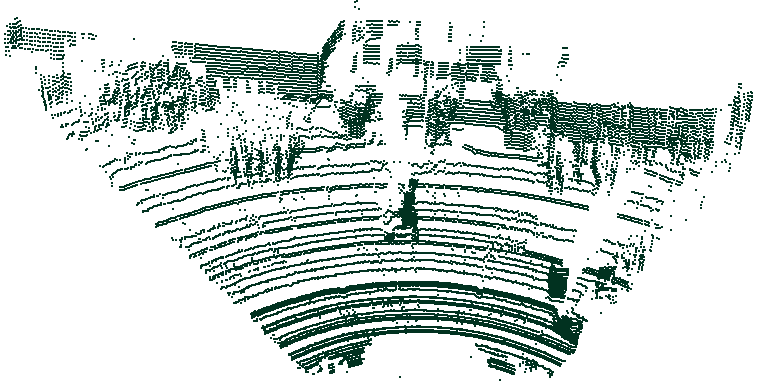}} \hfill
    \subfloat[Ours]{\includegraphics[width=  0.33\textwidth, height= 0.200\textwidth]{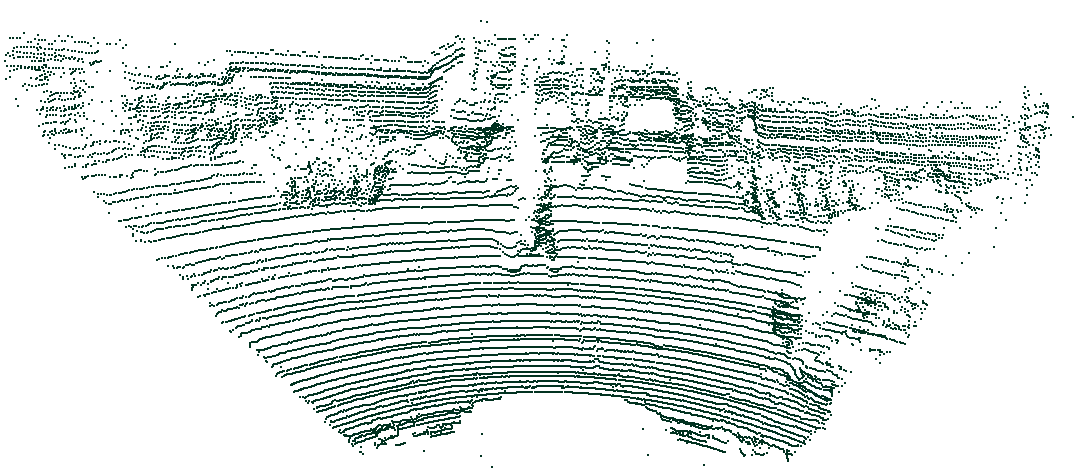}} \hfill
    \subfloat[Groundtruth]{\includegraphics[width=  0.33\textwidth, height=   0.200\textwidth]{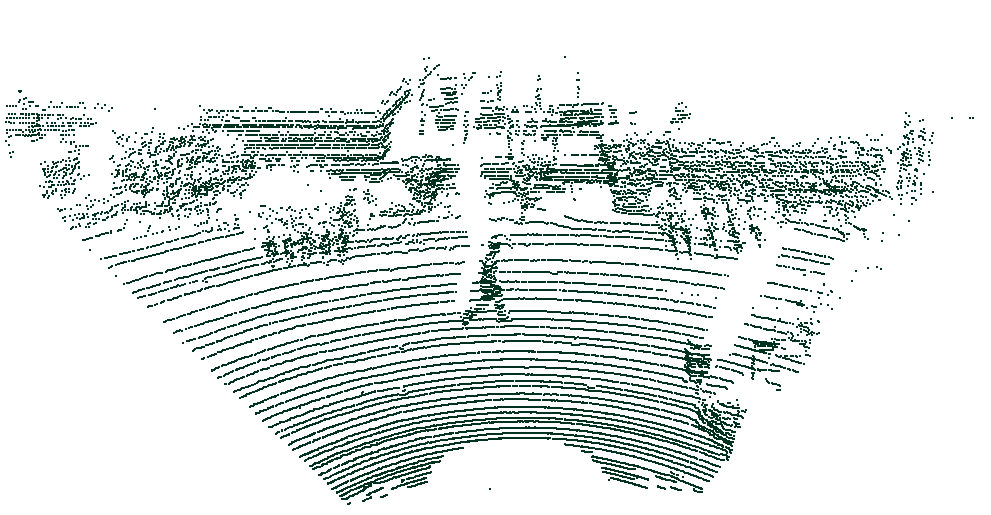}} 
    \end{minipage}
    
    \vspace{0.4cm}
    
    \begin{minipage}{.80\textwidth}
    \subfloat[Input]{\includegraphics[width=  0.33\textwidth, height=   0.200\textwidth]{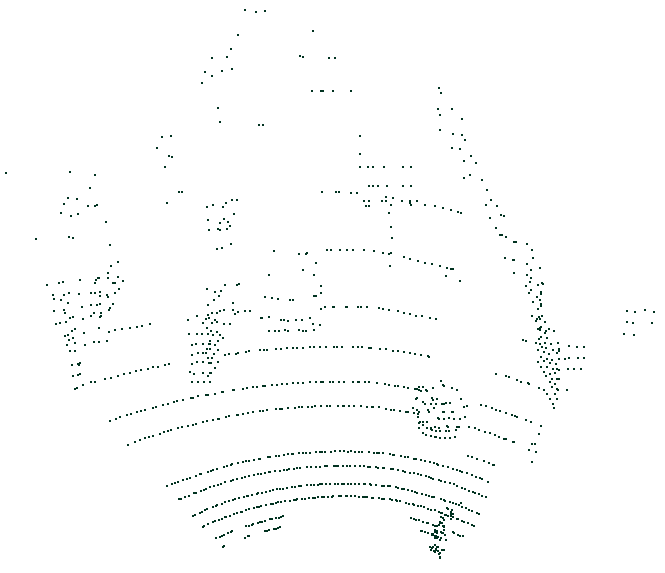}} \hfill
    \subfloat[LIDAR-SR\cite{unetlidar}]{\includegraphics[width=  0.33\textwidth, height=   0.200\textwidth]{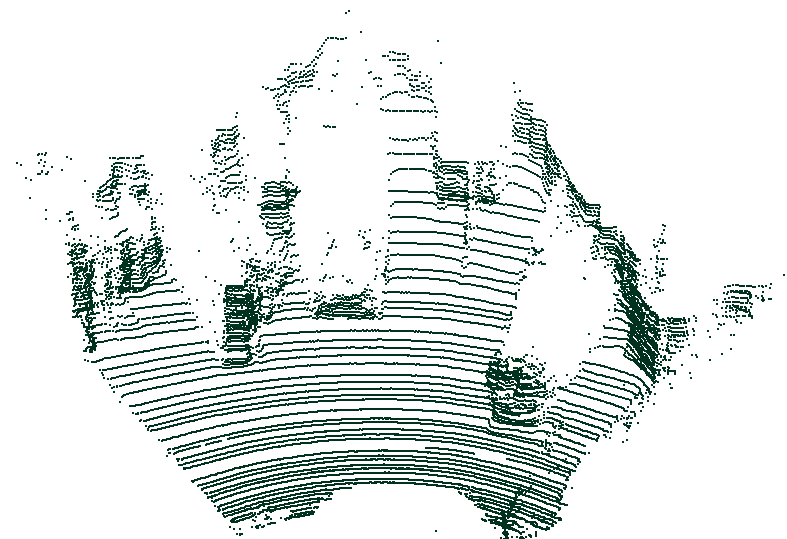}} \hfill
    \subfloat[SWIN-IR\cite{swinir}]{\includegraphics[width=  0.33\textwidth, height=   0.200\textwidth]{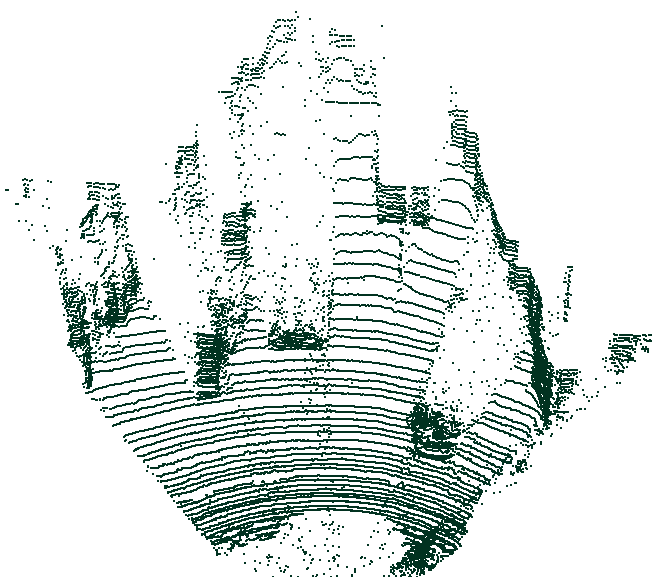}} \vfill
    \subfloat[ILN~\cite{iln}]{\includegraphics[width=  0.33\textwidth, height=   0.200\textwidth]{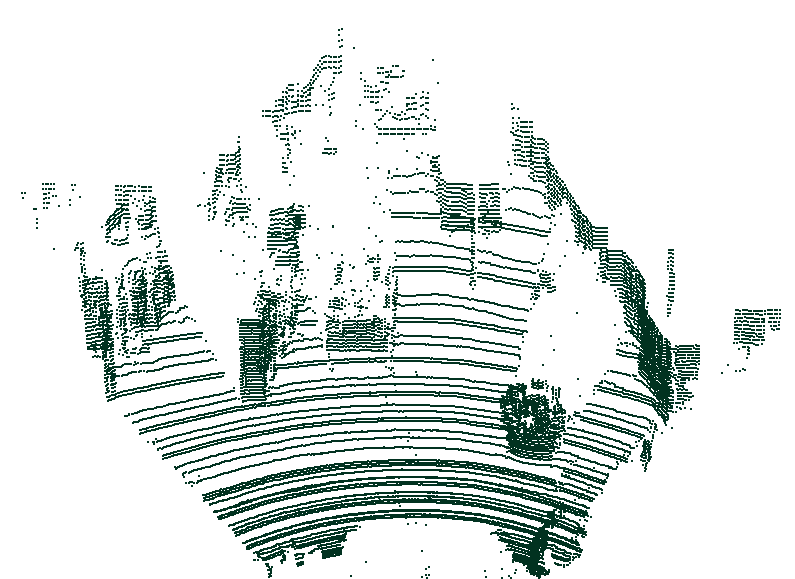}} \hfill
    \subfloat[Ours]{\includegraphics[width=  0.33\textwidth, height=   0.200\textwidth]{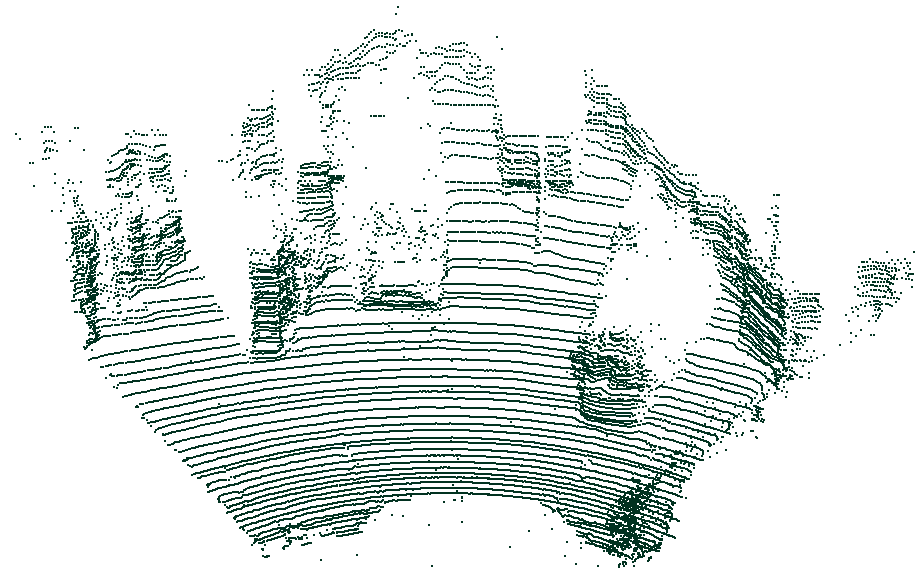}} \hfill
    \subfloat[Groundtruth]{\includegraphics[width=  0.33\textwidth, height=   0.200\textwidth]{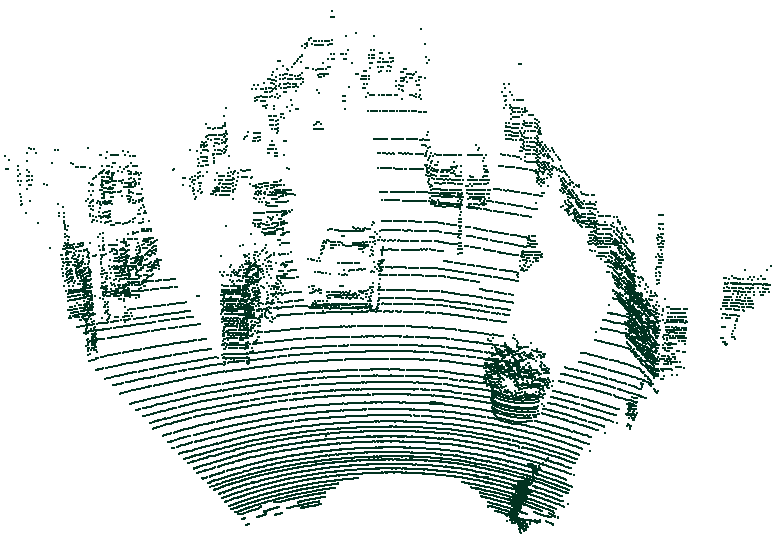}} 
    \end{minipage}
    
    \vspace{0.4cm}
    
    \begin{minipage}{.80\textwidth}
    \subfloat[Input]{\includegraphics[width=  0.33\textwidth, height=   0.200\textwidth]{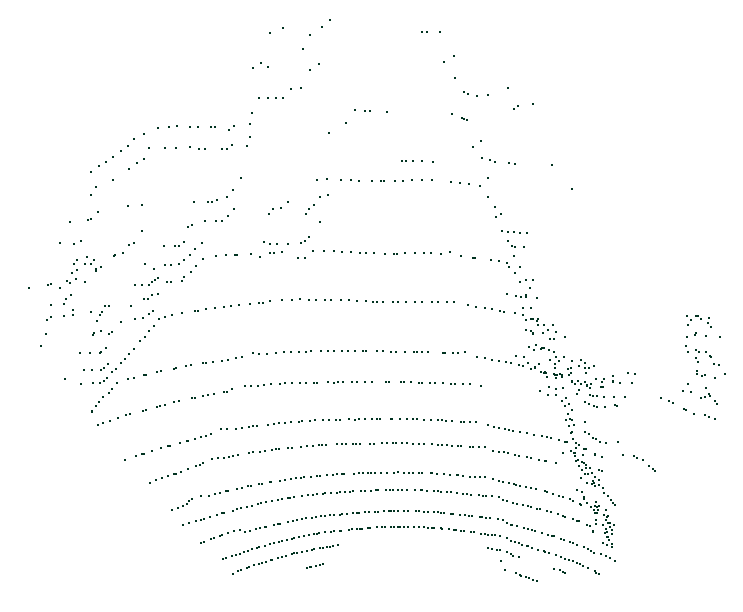}} \hfill
    \subfloat[LIDAR-SR\cite{unetlidar}]{\includegraphics[width=  0.30\textwidth, height=   0.200\textwidth]{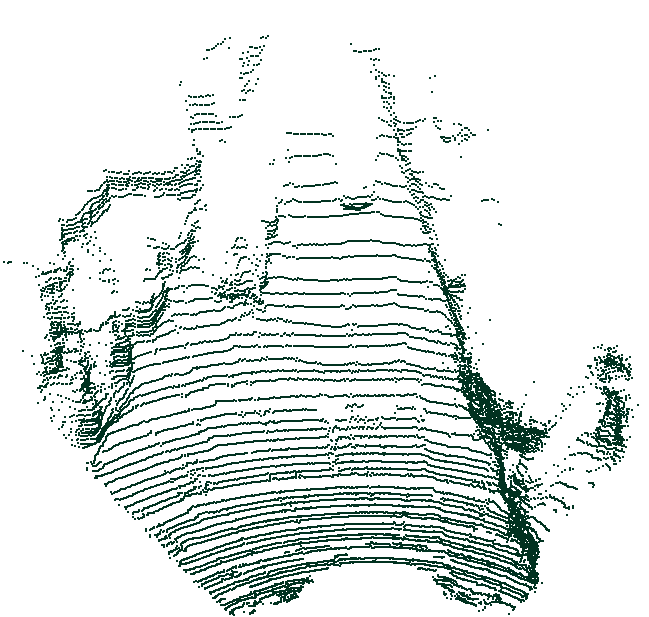}} \hfill
    \subfloat[SWIN-IR\cite{swinir}]{\includegraphics[width=  0.31\textwidth, height=   0.200\textwidth]{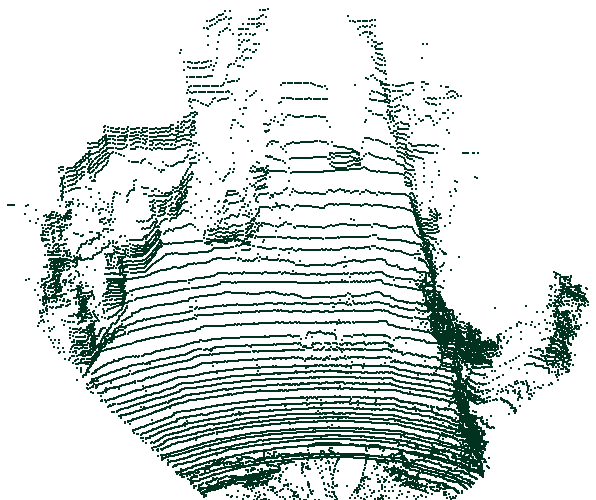}} \vfill
    \subfloat[ILN~\cite{iln}]{\includegraphics[width=  0.33\textwidth, height=   0.200\textwidth]{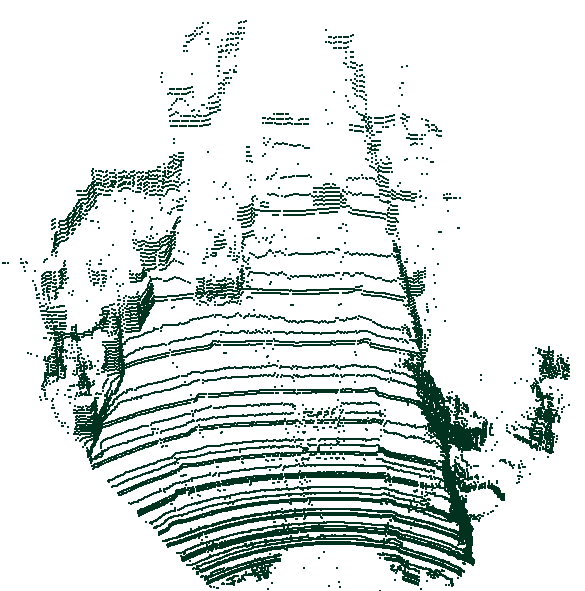}} \hfill
    \subfloat[Ours]{\includegraphics[width=  0.30\textwidth, height=   0.200\textwidth]{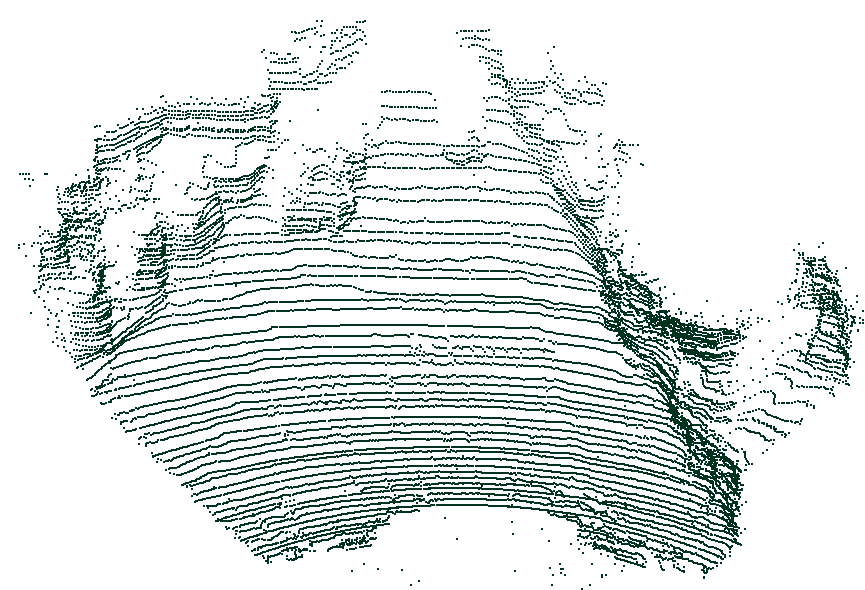}} \hfill
    \subfloat[Groundtruth]{\includegraphics[width=  0.33\textwidth, height=   0.200\textwidth]{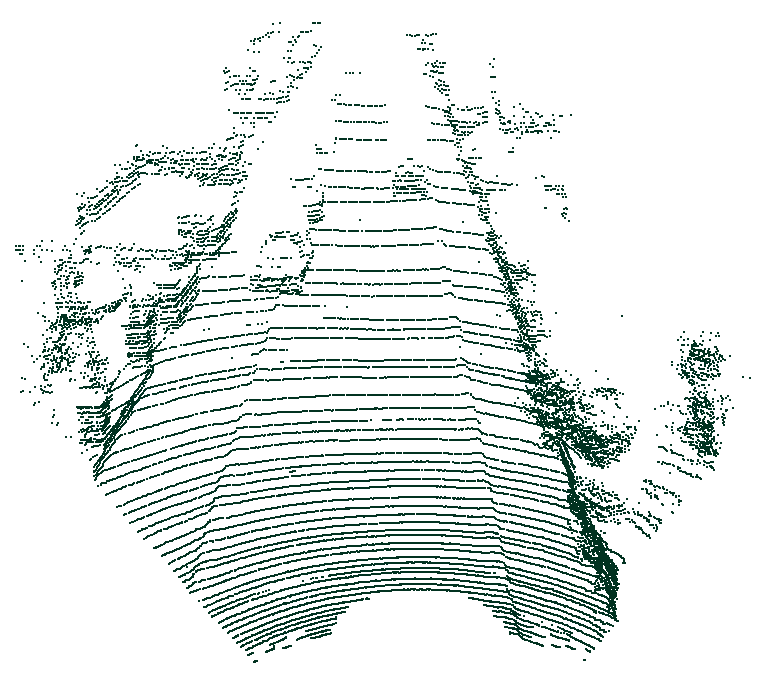}} 
    \end{minipage}

   \caption{Qualitative comparison on Cityscapes dataset}
   \label{fig:qualitative_scenes}
\end{figure*}